\documentclass[12pt,reqno]{amsart} 
\usepackage{graphicx,amssymb,colordvi,textcomp,latexsym,mathtools,enumitem}
\usepackage[hidelinks]{hyperref}
\usepackage{multirow,comment}
\usepackage{tabularx}
\usepackage{booktabs}
\usepackage{float}
\usepackage{prettyref}
\usepackage{chngcntr}
\counterwithout{equation}{subsection}
\counterwithout{equation}{section} 
\usepackage{lscape}
\usepackage{xcolor}

\def\ddefloop#1{\ifx\ddefloop#1\else\ddef{#1}\expandafter\ddefloop\fi}
\def\ddef#1{\expandafter\def\csname 
bb#1\endcsname{\ensuremath{\mathbb{#1}}}}
\ddefloop ABCDEFGHIJKLMNOPQRSTUVWXYZ\ddefloop
\def\ddef#1{\expandafter\def\csname 
	#1\endcsname{\ensuremath{\mathbf{#1}}}}
\ddefloop ABCDEFGHIJKLMNOPQRSTUVWXYZ\ddefloop
\def\ddef#1{\expandafter\def\csname 
#1\endcsname{\ensuremath{\mathbf{#1}}}}
\ddefloop abcdefghijklmnopqrstuvwxyz\ddefloop
\def\ddefloop#1{\ifx\ddefloop#1\else\ddef{#1}\expandafter\ddefloop\fi}
\def\ddef#1{\expandafter\def\csname 
b#1\endcsname{\ensuremath{\mathbb{#1}}}}
\ddefloop ABCDEFGHIJKLMNOPQRSTUVWXYZ\ddefloop
\def\ddef#1{\expandafter\def\csname 
c#1\endcsname{\ensuremath{\mathcal{#1}}}}
\ddefloop ABCDEFGHIJKLMNOPQRSTUVWXYZ\ddefloop
\def\ddef#1{\expandafter\def\csname 
f#1\endcsname{\ensuremath{\mathfrak{#1}}}}
\ddefloop ABCDEFGHIJKLMNOPQRSTUVWXYZabcdefghjklmnopqrstuvwxyz\ddefloop
\def\ddef#1{\expandafter\def\csname 
h#1\endcsname{\ensuremath{\widehat{#1}}}}
\ddefloop ABCDEFGHIJKLMNOPQRSTUVWXYZ\ddefloop
\def\ddef#1{\expandafter\def\csname 
hc#1\endcsname{\ensuremath{\widehat{\mathcal{#1}}}}}
\ddefloop ABCDEFGHIJKLMNOPQRSTUVWXYZ\ddefloop

\newcommand{\balpha}{\boldsymbol{\alpha}}

\newcommand{\bbeta}{\boldsymbol{\beta}}

\newtheorem{lemma}{Lemma}[section]

\newtheorem{prop}[lemma]{Proposition}
\newtheorem{thm}[lemma]{Theorem}
\newtheorem*{thm*}{Theorem}
\newtheorem{cor}[lemma]{Corollary}
\theoremstyle{definition}
\newtheorem{definition}[lemma]{Definition}
\newtheorem{exam}[lemma]{Example}

\theoremstyle{remark}

\newtheorem{rem}[lemma]{Remark}
\newtheorem{rems}[lemma]{Remarks}

\newcommand{\bones}{\mathbf{1}}

\newcommand{\defeq}{\coloneqq}
\DeclarePairedDelimiter{\brk}{[}{]}
\DeclarePairedDelimiter{\crl}{\{}{\}}
\DeclarePairedDelimiter{\prn}{(}{)}
\DeclarePairedDelimiter{\nrm}{\|}{\|}
\DeclarePairedDelimiter{\inner}{\langle}{\rangle}
\DeclarePairedDelimiter{\tri}{\langle}{\rangle}

\DeclarePairedDelimiter{\floor}{\lfloor}{\rfloor}

\DeclareMathOperator*{\argmin}{arg\,min} % * Places subscript 
\newcommand{\minimizers}{X_\c^m}
\newcommand{\symp}{\mathrm{Sym}}

\newcommand{\reals}{\mathbb{R}}

\newcommand{\sphere}{\cS}
\newcommand{\ball}{\cB}
\newcommand{\cl}[1]{\overline{#1}}

\usepackage{prettyref}
\newcommand{\pref}[1]{\prettyref{#1}}

\newcommand{\savehyperref}[2]{\texorpdfstring{\hyperref[#1]{#2}}{#2}}
\newrefformat{eq}{\savehyperref{#1}{\textup{(\ref*{#1})}}}
\newrefformat{eqn}{\savehyperref{#1}{Equation~\ref*{#1}}}
\newrefformat{lem}{\savehyperref{#1}{Lemma~\ref*{#1}}}
\newrefformat{obs}{\savehyperref{#1}{Observation~\ref*{#1}}}
\newrefformat{def}{\savehyperref{#1}{Definition~\ref*{#1}}}
\newrefformat{thm}{\savehyperref{#1}{Theorem~\ref*{#1}}}
\newrefformat{cor}{\savehyperref{#1}{Corollary~\ref*{#1}}}
\newrefformat{sec}{\savehyperref{#1}{Section~\ref*{#1}}}
\newrefformat{app}{\savehyperref{#1}{Appendix~\ref*{#1}}}
\newrefformat{fig}{\savehyperref{#1}{Figure~\ref*{#1}}}
\newrefformat{alg}{\savehyperref{#1}{Algorithm~\ref*{#1}}}
\newrefformat{rem}{\savehyperref{#1}{Remark~\ref*{#1}}}
\newrefformat{rems}{\savehyperref{#1}{Remarks~\ref*{#1}}}
\newrefformat{conj}{\savehyperref{#1}{Conjecture~\ref*{#1}}}
\newrefformat{prop}{\savehyperref{#1}{Proposition~\ref*{#1}}}
\newrefformat{table}{\savehyperref{#1}{Table~\ref*{#1}}}
\newrefformat{prob}{\savehyperref{#1}{Problem~\ref*{#1}}}
\newrefformat{claim}{\savehyperref{#1}{Claim~\ref*{#1}}}
\newrefformat{fact}{\savehyperref{#1}{Fact~\ref*{#1}}}
\newrefformat{sec}{\savehyperref{#1}{Section~\ref*{#1}}}
\newrefformat{exam}{\savehyperref{#1}{Example~\ref*{#1}}}
\newrefformat{prob}{\savehyperref{#1}{(\ref*{#1})}}

\usepackage{setspace}

\newcommand{\RR}{\mathbb{R}}
\newcommand{\EE}{\mathbb{E}}
\newcommand{\QQ}{\mathbb{Q}}
\newcommand{\PP}{\mathbb{P}}
\newcommand{\CC}{\mathbb{C}}
\newcommand{\ZZ}{\mathbb{Z}}
\newcommand{\NN}{\mathbb{N}}

\newcommand{\rk}{{\rm rk}}
\newcommand{\ploss}{{\cL}} 
\renewcommand{\ker}{{\kappa}}
\newcommand{\codim}{\text{codim}}

\newcommand{\bxi}{{\boldsymbol{\xi}}}

\newcommand{\grad}[1]{{\nabla {#1}}}

\newcommand{\is}[1]{{[#1]}}

\newcommand{\dd}{\mbox{$\;|\;$}}
\newcommand{\trans}{\pitchfork}

\newcommand{\im}{\mathrm{im}}
\newcommand{\lker}{\mathrm{ker}}
\newcommand{\tr}{\mathrm{Tr}}

\renewcommand{\hom}{\mathrm{Hom}}

\newcommand{\GL}[1]{{\text{\rm GL}({#1})}}
\newcommand{\defoo}{~\dot{=}~}

\newcommand{\arr}{ \to }

\usepackage{breqn}

\newcommand{\real}{\mathbb{R}}

\newcommand{\critset}{\Sigma}
\newcommand{\tanset}{\mho}
\newcommand{\gspace}[1]{\cC^{\infty}_{#1}}
\newcommand{\fspace}[2]{C^\infty_{#1}\prn{#2}_0}
\newcommand{\fspacec}[3]{C^\infty_{#1}\prn{#2}_{#3}}

\newcommand{\masterset}{Z}
\newcommand{\masterstrat}{\cZ}
\newcommand{\ibr}[1]{[#1]}
\newcommand{\propC}{{Property $(\fR_{1,2})$}}
\newcommand{\suf}{$\chi$-}
\newcommand{\uniset}{\cU}
\newcommand{\secsymres}{Section~6}
\newcommand{\secsuf}{Section~7}
\newcommand{\seccon}{Section~8}
\newcommand{\closure}[1]{\overline{#1}}

\title{Symmetry \& Critical Points}

\author{Yossi Arjevani}
\email{yossi.arjevani@gmail.com}

\begin{document}
\begin{abstract}
Critical points of an invariant function may or may not be symmetric. We prove, however, that if a symmetric critical point exists, those adjacent to it are generically symmetry breaking. This mathematical mechanism is shown to carry important implications for our ability to efficiently minimize invariant nonconvex functions, in~particular those associated with neural~networks.

\end{abstract}
\maketitle
\tableofcontents

% !TEX root = general_symmetry_and_critical_points.tex

\section{Introduction}
Suppose given a finite group\footnote{Most of what we say applies to compact Lie group, see \cite{arjevani2025optsym}.} $G$ acting on a vector space $V$, and let $f:V\to\RR$ be a smooth $G$-invariant function. The largest subgroup $G_\x \le G$ fixing a point $\x\in V$  is called the \emph{isotropy} group and is generally used as a means of measuring symmetry. If $f$ is strictly convex, we may average over $G$ and so find that, if attained, the minimum $\c$ must be fixed by all group elements, giving $G_{\c}=G$. For nonconvex functions, however, symmetry of minima and critical points may break arbitrarily, allowing in particular for trivial isotropy groups. In this article, we develop methods to rigorously establish the occurrence of symmetry breaking (SB) critical points having \emph{large} isotropy groups. 
Granted the existence of a symmetric critical point $\c\in V$, the approach proceeds by constructing arcs that emanate from $\c$ and inherit some of its symmetries. A critical point hit by the arc must then be at least as symmetric as the arc; hence~SB. The significance of such phenomena, as demonstrated in the paper, lies in the fact that, when present, SB makes possible the use of powerful methods for studying the tractability of nonconvex problems and allows for an efficient organization of an otherwise highly complex set of critical points.

In general terms, this article developed out of a program to understand why the highly nonconvex loss landscapes involved in the training of neural networks on natural distributions can be efficiently minimized using local optimization methods. However, the determination of symmetry and stability of critical points is an issue central to several scientific areas. The paper is thus intended for a general machine learning and mathematical optimization readership, as well as theoretical physicists interested in the theory of neural networks. Our presentation has been organized accordingly to include substantial preliminary sections aimed at introducing the relevant concepts as necessary.

% !TEX root = general_symmetry_and_critical_points.tex

\section{Main results: symmetry breaking critical points}
A formal discussion of our results requires some familiarity with  differential topology. We defer detailed statements to later sections after the relevant notions have been introduced and, for now, emphasize a high-level informal overview of the results: those related to the mathematical mechanism accounting for the occurrence of SB critical points are presented in this section, those pertaining to the use of SB in the study of invariant nonconvex  problems in \pref{sec:apps}. 

The general strategy indicated earlier for relating the symmetry of a critical point $\c$ to those adjacent to it is implemented by constructing arcs that describe points where the level sets of a given function lie \emph{tangential} to $\sphere_\c(r)$, the radius-$r$ sphere centered at $\c$.

\begin{definition}[Tangency sets and arcs]\label{def:tanset}\phantom{}Let $f:\RR^d\to\RR$ be a $C^1$ function.
\begin{enumerate}[leftmargin=*]
\item The set of critical points of $f$ is denoted by $$\critset(f) \defeq \{\x\in \RR^d~|~D f(\x) = 0\}.$$ 
\item The \emph{tangency set} of $f$ relative to  $\c\in\RR^d$ is defined by 
\begin{align*}
\tanset_\c(f) \defeq \{ \x \in\RR^d~|~ D_if(\x)(\x-\c)_j  = D_jf(\x)(\x-\c)_i, i,j\in[d]  \}.
\end{align*}
In particular, $\critset(f)\subseteq \tanset_\c(f)$.  
\item
A \emph{tangency arc} relative to $\c\in\RR^d$ is a continuous curve $\gamma:[0,\epsilon)\to \RR^d,~\epsilon>0,$ satisfying $\gamma(0) = \c$ and $\gamma(t) \in \tanset_\c(f)\backslash\{\c\}$ otherwise.
\end{enumerate}
\end{definition}
By a simple Lagrangian multipliers argument, any minimizer of $f|_{\sphere_\c(r)}$ satisfies the condition stated in (2) above, and so by compactness the tangency set is never empty. The topology of the tangency set, however, can be quite complicated for arbitrary $C^1$ functions. We begin by considering homogeneous quadratic polynomials, where the tangency set has a particularly simple form. If $f:\RR^d\to\RR$ is given by  $f(\x) = \x^\top A \x/2$, $A$ a symmetric matrix, then relative to $\c=0$,
\begin{equation}\label{eqn:quad_tan}
	\tanset_0(f) = \{\x \in \RR^d~|~\exists \eta\in\RR, A\x= \eta \x\},
\end{equation}
comprising the origin and all Hessian eigenvectors. This simple example illustrates how the tangency set may be locally constrained by the derivatives of $f$, here the spectral properties of the Hessian (assuming now $C^2$). Further information on the tangency set may and shall be extracted by accounting for constraints involving higher-order derivatives, collectively referred to as \emph{tangency equations}, see \pref{def: tan_eqns}. For example, if $k\ge 3$ and $p$ is a $k$-degree homogeneous polynomial, then $\mho_0(p)\backslash\{0\}$ is the set of all tensor eigenvectors (specifically, non-normalized E-eigenvectors \cite{qi2005eigenvalues}) of $D^kp(0)$.

Referring to (\ref{eqn:quad_tan}), if $A= I_d$, then $\tanset_0(f) =\RR^d$ and so despite $\c = 0$ being highly symmetric (fixed by any linear group action), points of trivial isotropy exist in the tangency set (abundantly so). This might seem discouraging in view of our general plan to argue about the structure of critical points via the tangency set. However, the situation for $\tanset_0(f)$ is unstable in the sense that in any arbitrarily small neighborhood of~$f$, almost every function has exactly $2d$ Hessian eigenvectors of norm one. The latter statement presupposes a choice of topology on a suitable function space. We defer formal definitions to \pref{sec: tan pre} and assume for now an intuitive notion of a function's neighborhood. An assertion holding for an open and dense set with respect to the chosen topology is said to be \emph{generic}. 

The example described illustrates our general approach,  drawing on ideas originating from the work of Whitney, Thom and Mather: we restrict to a class of functions that facilitates a complete classification of all tangency sets---yet generic. The approach allows us to provide detailed results that are characteristic of functions invariant under a given group action and are robust under perturbations. We carry out this program for a number of group representations, emphasizing permutation representations, see \pref{sec:pre_sym}. In the absence of symmetry, we have the following general result.
\begin{thm*}[Informal,  no symmetry] \label{thm: class_informal_nosym} 
Generically,  the tangency set of a smooth function on $\RR^d$ consists of $2d$ tangency arcs, each tangential to a Hessian eigenspace.
\end{thm*}
\begin{figure}[t]
	\includegraphics[scale=0.5]{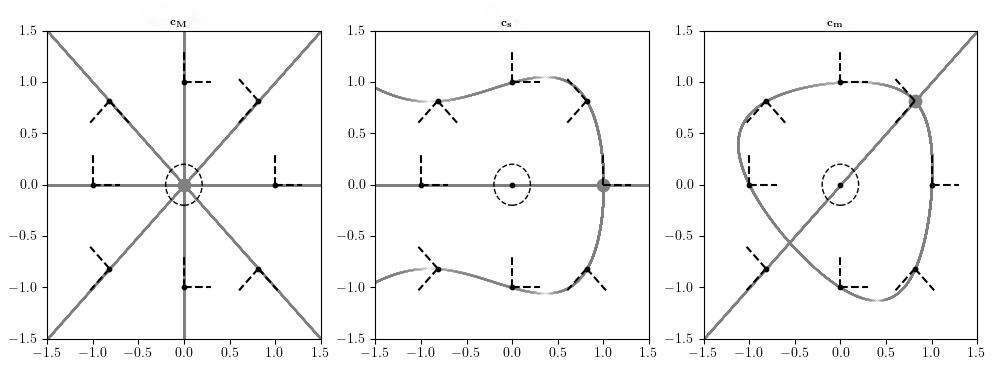}
	\caption{Tangency sets of $h$ defined in (\ref{eq:h}) relative to different critical points $\c_{\bullet}$. The tangency arcs \emph{break} the symmetry of $\c_{\bullet}$ to a varying degree.	Critical points connected by tangency arcs are at least as symmetric as the arcs and so are themselves symmetry breaking. \label{fig:hc_tan}}
\end{figure}
To demonstrate the type of results holding under symmetry, consider the following simple $B_2$-invariant function, $B_n$ generally denoting the hyperoctahedral group (Weyl group of type~$B$),
\begin{equation}\label{eq:h}
	h(x,y) = x^4 + x^2y^2 + y^4 -2x^2 - 2y^2.
\end{equation}

\begin{figure}[h]\label{fig: sche}
\includegraphics[scale=0.35]{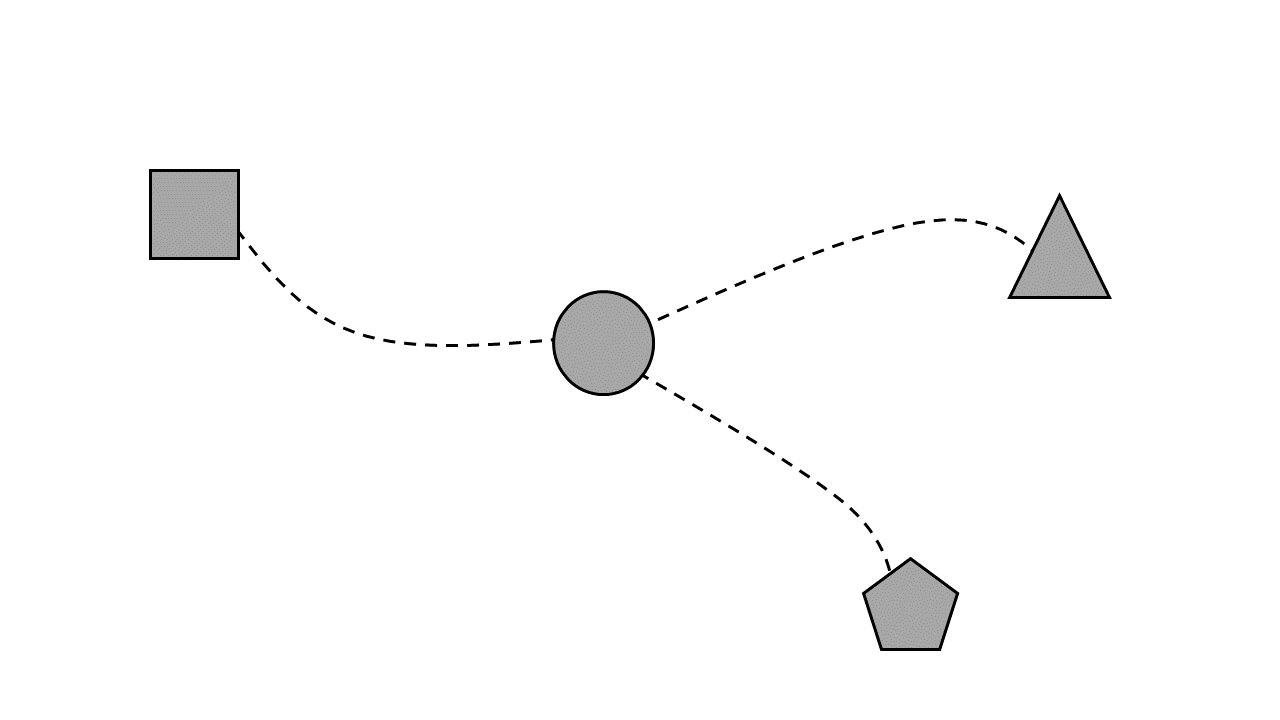}
\caption{Critical points connected through tangency arcs to a symmetric point, here a circle, break the symmetry by a process resembling (simultaneous) spontaneous symmetry breaking.}
\end{figure}

The function $h$ is invariant under $(x,y) \mapsto (y,x)$, $x\mapsto-x$ and $y\mapsto-y$,  hence $B_2$-invariant. Computing, we find that $h$ has nine critical points: $\c_M = (0,0)$ (a maximum), $\c_s = (1,0)$ giving four critical points (saddles) in total owing to its \emph{orbit} $B_2(1,0) = \{ g(1,0)~|~g\in B_2\}$, and $\c_m = (\sqrt{3/2}, \sqrt{3/2})$ (a minimum) again yielding $|B_2(\sqrt{3/2}, \sqrt{3/2})|=4$ critical points. By equivariance, it suffices to compute the tangency set for the orbit representatives: $\c_M, \c_s$ and $\c_m$. Although in this case the tangency set can be described by simple algebraic expressions, our interest lies rather in the geometric situation. Referring to \pref{fig:hc_tan}, the isotropy groups of tangency arcs are subgroups of $(B_2)_{\c_{\bullet}}$---arcs having nontrivial isotropy give a lower bound on the isotropy of the critical points to which they~connect (see also \pref{fig: sche}). Additional properties holding in the general situation are as follows.\\

\noindent\textbf{Proposition}  (informal) \label{prop:omin} Generically, \\
(1) A tangency set consists of a finite number of connected one-dimensional manifolds that share a common boundary at the center $\c$.\\
(2) Any critical point lying sufficiently close to $\c$ may be connected by a tangency arc.\\

Here, $\tanset_{\c_M}(h)\backslash\{\c_M\}$ is a one-dimensional manifold having eight connected components (indeed, arcs) connecting to all critical~points of~$h$. 

If $f$ is a smooth function and $\c\in\RR^d$, we denote by $T_\c^k(f)$ the Taylor polynomial of degree $k$ at $\c$. Given the detailed information on $\tanset_{\c_M}(h)$, it is natural to ask whether the tangency set of a smooth function $f$ satisfying $T^4_{0}(f) = h$ can be related to the former in terms of structure and symmetry. The general investigation of questions of this nature originated from the study of singularities of maps. Of the several equivalence relations available in singularity theory, a natural choice in our context is the \emph{isotopic} equivalence and its extension to the equivariant category \cite{bierstone1976generic}. As earlier, we defer formal definitions to \pref{sec: tan pre}, and content ourselves with an intuitive use of the concept. Thus, two tangency sets are equivalent if one can be continuously deformed into the other---equivariantly so. A~smooth function is said to be finitely determined if its tangency set is equivalent to one associated with one of its Taylor polynomials. We typically do not pursue the minimal such degree. \\

\noindent\textbf{Proposition}. If $(V,G)$ is a permutation representation, then generically $G$-invariant functions are finitely determined.\\

The result follows by a well-known bound on the degrees of a minimal set of generators for the invariants of permutation groups (\cite{garsia1984group}, and see \cite[Theorem 6.2.9]{smith1995polynomial} for a more direct homology theoretic proof).

Reducing the characterization of a tangency set to that of a finite model given by the Taylor polynomial, we may now determine the \emph{admissible} isotropy groups persisting under perturbations. The following holds for the defining representation $(\RR^d, S_d)$ of $S_d$, the symmetric group on $d$ symbols.
\begin{thm*}(informal). Generically, $(\RR^d, S_d)$-invariant functions are finitely determined, the number of tangency arcs is at most exponential in the dimension, and the admissible isotropy groups occurring for $(\RR^d, S_d)$-tangency sets are conjugated to $S_{d-q}\times S_q$, $q=0,\dots, \floor{\frac{d}{2}}$.
\end{thm*}
The result is related to ones concerning the irreducible standard representation of $S_d$, see \pref{sec: tan sd}. A notable distinction is that here a tangency arc exhibiting the same isotropy as $\c$ may indeed exist and connect to other critical points, as seen for example by $\tanset_{0}((x,y)\mapsto 3x^2 -2xy + 3y^2 \allowbreak+ x^3+ y^3)$. 

Additional permutation representations central to this work are $(M(k,\allowbreak d),S_k\times S_d)$, given by the external tensor product of $(\RR^k, S_k)$ and $(\RR^d, S_d)$, and the usual tensor product $(M(k,d), S_d)$ with $S_d$ acting diagonally ($k\ge d$). We henceforth assume $k=d$ and defer the case where $k$ and $d$ are decoupled to
\cite[\secsymres]{arjevani2024sc2}. The significance of the representation $(M(d,d), S_d)$ lies in its encoding of symmetries found in a fully connected two-layer ReLU network. Define
\begin{align} \label{net}
N(\x; W, \a) \defeq
\a^\top\!\sigma\prn{W\x},\quad
~W\in  M(d,d),~\a \in \RR^d,
\end{align}
where $\sigma$ is the ReLU activation function $\max\crl{0,x}$ acting entrywise and $M(d,d)$ is the space of $d\times d$ matrices. If data is fully realizable, the nonconvex optimization problem associated with the squared loss is given by 
\begin{align}\label{opt:problem_relu}
\cL(W, \balpha) \defeq \frac{1}{2}\EE_{\x\sim 
\cP}\brk*{\Big(f(\x; W, \a) -      f(\x; T, 
\boldsymbol{b})\Big)^2},
\end{align}
where $\cP$ denotes a probability distribution on the input space, $W\in M(d,d), \a \in \RR^d$ denote the optimization variables, and $T\in M(d,d), \b \in \RR^d$ are fixed parameters \cite{simsek2024should, safran2022effective,arjevani2021analytic} (questions involving the empirical loss defined for a finite number samples are considered in \cite{arjevani2024dsb}). The invariance properties of $\ploss$ are determined by the choice of  $\cP$ and the target weights $T$ and $\b$. We start with a symmetric model and subsequently break symmetry. Assume that the second layer is fixed and set to $\bones_d$. If, by way of demonstration, $\cP$ is the $d$-variate normal distribution and $T=I_d$, then $(S_d)_T = \Delta S_d$, $\Delta G$ generally denoting the diagonal embedding of a group $G$ in $G\times G$, and $\ploss$ is $(M(d,d),S_d)$-invariant.

\begin{thm*}(informal). Generically, $(M(d,d), S_d)$-invariant functions are finitely determined, the number of tangency arcs is exponential in the dimension, and  some of the admissible isotropy groups occurring for $(M(d,d), S_d)$-tangency sets are $\Delta S_d, \Delta (S_{d-1}\times S_1), \Delta (S_{d-2}\times S_1^2),  \Delta (S_{d-2}\times \ZZ_2), \Delta (S_{d-3}\times \ZZ_3)$ and $\Delta (S_{d-4}\times \ZZ_4)$ (modulo conjugation).
\end{thm*}
A comprehensive list of all admissible isotropy groups is given in \secsymres~after the relevant group-theoretic notions have been introduced. We also elaborate in that section on how the results apply to subgroups of $S_d$, e.g., of the form  $(M(d,d), S_{d-p}\times S_p)$, and to target weight matrices with smaller isotropy groups.

The admissible isotropy groups predicted by our theory align with those observed empirically for two-layer ReLU networks \cite{arjevanifield2019spurious}. Of course, there is nothing special about the ReLU activation---the results hold generically for $(M(d,d), S_d)$-invariant functions. An additional fundamental setting demonstrating the explanatory power of the approach is that of tensor decomposition problems \cite{arjevani2021symmetry}. As in the case of ReLU, certain \emph{identifiability} also applies in the latter in the sense that the set of global minima is precisely the $S_d$-orbit of $T$. In particular, all global minima have isotropy $\Delta S_d$ and so are highly symmetric. Whenever this holds, SB phenomena are to be expected, just as in various instances of the following more general formulation,
\begin{align}\label{opt problem}
\ploss_\ker&(W,\balpha;V,\bbeta) \defeq\\
&\sum_{i,j=1}^{k} \alpha_i\alpha_j\ker\prn{\w_i, \w_j}
-2\sum_{i=1}^{k} \sum_{j=1}^{h} \alpha_i\beta_j\ker\prn{\w_i, \v_j}
+ \sum_{i,j=1}^{h}\alpha_i\beta_j \ker\prn{\v_i, \v_j},\nonumber
\end{align}
with $W\in M(k,d),V\in M(h,d)$, $\alpha_i,\beta_i$ real scalars, and $\kappa:\RR^d\times\RR^d\to \RR$ a \emph{kernel} function. Note that contrary to the conventional use of kernels, in (\ref{opt problem}),  both $\alpha_i$ and $\w_i$ are trainable. Moreover, for the purpose of establishing the invariance properties of $\ploss$, $\ker$ is not required to be positive definite. 
A rich class of kernels, subsuming the soft committee machine studied within the framework of statistical mechanics \cite{saad1995line,goldt2019dynamics} and the so-called multilayer kernel machines \cite{ChooSaul2009}, is given by
\begin{align} \label{ker_cd}
\ker_\cP(\w,\v) \defoo \bE_{\x \sim \cP} [\rho(\inner{\w,\x})\rho(\inner{\v,\x})],
\end{align}
where $\cP$ as earlier denotes a distribution over $\RR^d$ and $\rho:\reals\to\reals$ is integrable. Choosing the ReLU activation function and $\cP=\cN(0,I_d)$ recovers \cite{arjevanifield2019spurious}. Replacing the ReLU activation  by $\rho(\xi) = \xi^3$ yields $\ker_\cN(\w,\v) = {6}\inner{\w,\v}_F^3 + 9\|\w\|_2^2\|\v\|_2^2\inner{\v,\w}_F$ giving the symmetric tensor decomposition problem under certain tensor norm. The Frobenius tensor norm is recovered by setting $\kappa(\w,\v) = \inner{\w,\v}^n$, giving
\begin{align} \label{prob:ext}
\ploss(W,\balpha;V,\bbeta) = \nrm*{ \sum_{i=1}^{k}\alpha_i 
	\w_i^{\otimes n}- \sum_{i=1}^{h}\beta_i 
	\v_i^{\otimes n}}^2.
\end{align}
The above examples may be generalized by the following formulation,
\begin{align}\label{geo ker}
\ker(\w,\v) = \|\w\|^k\|\v\|^kJ(\cos(\w,\v)),
\end{align}
for a suitable univariate function $J$. We study (\ref{geo ker}) in some depth in~\secsymres.

We conclude this section with a general remark following  \cite[Section 5]{arjevani2023hidden}. If we may engage in some wild speculation, let us assume that nature generates labels by an unknown neural network yielding symmetric global solutions for a choice of loss function. The mathematical mechanism introduced in this work then suggests that the optimization landscape associated with the training of a neural network would generically exhibit SB critical points. SB implies strict structural constraints on critical points and dynamics (see  \pref{sec:apps}), in particular ones that turn minima into saddles as the number of parameters increases, thereby allowing gradient-based methods to detect models that generalize~well.

% !TEX root = general_symmetry_and_critical_points.tex

\section{Applications} \label{sec:apps}
We present several applications demonstrating the use of SB in studying invariant nonconvex optimization problems. Some of the applications have appeared in previous work; others are novel. Throughout, we work over standard Euclidean spaces.

\subsection{Hessian, Gauss-Newton and group representation theory} \label{sec: app rep}
SB simplifies, in many cases enables, an explicit computation of the Hessian spectrum by making possible the use of methods from group representation theory, see e.g., \cite[Section XII]{golubitsky2012singularities} and \pref{sec: reps}.

Given two representations $(V,G)$ and $(W,G)$, a map $A: V \arr W$ is called $G$-equivariant if $A(gv) = gA(v)$, for all $g \in G, v \in V$. If $A$ is linear and equivariant, we say that $A$ is a \emph{$G$-map}. Invariant functions $f:V\to \RR$ naturally provide examples of equivariant maps. Thus the gradient $\nabla f$ is a $G$-equivariant self map of $\RR^d$, and if $\c$ is a critical point of $\nabla f$ then $\x\mapsto\nabla^2 f(\c)\x$ is a $G_\c$-map. The equivariance of the Hessian is the key ingredient that allows us to study the Hessian spectrum at SB critical points and local minima.

Every representation $(\real^n,G)$ can be written uniquely, up to order, as an orthogonal direct sum $\oplus_{i\in\ibr{m}} V_i$, where each $(V_i,G)$ is an orthogonal direct sum of isomorphic irreducible representations $(V_{ij},G)$, $j \in 
\ibr{p_i}$, and $(V_{ij},G)$ is isomorphic to $(V_{i'j'},G)$
if and only if $i' = i$. The subspaces $V_{ij}$ are \emph{not} uniquely determined if $p_i > 1$. If there are $m$ distinct isomorphism classes $\mathfrak{v}_1,\cdots,\mathfrak{v}_m$ of irreducible representations, then $(\real^n,G)$ may be represented by the sum $p_1 \mathfrak{v}_1 + \cdots + p_m \mathfrak{v}_m$, where $p_i\ge 1$ counts the number of representations with isomorphism class $\mathfrak{v}_i$. Up to order, this sum (that is, the $\mathfrak{v}_i$ and their multiplicities) is uniquely determined by $(\real^n,G)$. This is the \emph{isotypic decomposition} of $(\real^n,G)$ (see \cite{thomas2004representations} and \pref{sec: reps}).

The isotypic decomposition is a powerful tool for extracting information about the spectrum of $G$-maps. For example, take $G = S_d$.
Every irreducible representation of $S_d$ is real~\cite[Thm.~4.3]{fulton1991representation}. Suppose, as above, that 
$(\real^d,S_d)=\oplus_{i\in\ibr{m}} V_i$ and $A: \real^d\arr\real^d$ is an $S_d$-map. 
Since the induced maps $A_{ii'}: V_i \arr V_{i'}$ must be zero if $i \ne i'$, $A$ is uniquely determined by
the $S_d$-maps $A_{ii}: V_i \arr V_i$, $i \in\ibr{m}$. Fix $i$ and choose an $S_d$-representation $(W,S_d)$
in the isomorphism class $\mathfrak{v}_i$. Choose $S_d$-isomorphisms $W \arr V_{ij}$, $j \in \ibr{p_i}$.
Then $A_{ii}$ induces $\overline{A}_{ii}: W^{p_i} \arr W^{p_i}$ and so determines a (real) matrix $M_i \in M(p_i,p_i)$ since $\text{Hom}_{S_d}(W,W) \approx \real$. Different choices of $V_{ij}$, or isomorphism $W \arr V_{ij}$, yield a matrix similar to $M_i$. Each eigenvalue of $M_i$ of multiplicity $r$ gives an eigenvalue of $A_{ii}$, and so of $A$, of multiplicity 
$r\,\text{degree}(\mathfrak{v}_i)$. In particular, $A$ has most $\sum_{i \in \ibr{m}}p_i$ distinct real eigenvalues.
If the isotypic decomposition stabilizes for large $d$, as with the representations considered in \pref{sec: stab chars}, the bound holds independently of the dimension of the underlying space. 

Consider, for example, $(\RR^d, S_d)$. In the general theory, each irreducible representation $V_\lambda$ of $S_d$ is associated to a partition $\lambda$ of $d$. If $d\ge 2$, 
\begin{align}
\RR^d &= V_{(d)} \oplus V_{(d-1, 1)},\label{iso_decomp_sym1}
\end{align}
and so although an $(\RR^d, S_d)$-invariant map has $d$ Hessian eigenvalues, the number of distinct eigenvalues is at most 2. The technique was used in \cite{arjevani2023hidden,arjevani2023symmetry,arjevani2021analytic,arjevani2020hessian} to compute the Hessian spectrum of SB critical points, see \pref{table: skew}. An additional important example of analytic quantities amenable to the same group representation technique is provided by the Gauss-Newton decomposition where both terms are $G$-maps. 

\begin{figure}[h]
\begin{tabular}{l|l}\label{table: skew}
	Eigenvalue & Multiplicity\\\hline
	\rule{0pt}{4ex}
	$O(d^{-1/2})$ & $d$ \\
	\rule{0pt}{3ex}    
	$
	%	\frac{\pi -2}{4 \pi} 
	\frac{1}{4} - \frac{1}{2\pi}
	+ O(d^{-1/2})$ & $\frac{(d-1)(d-2)}{2}$ \\
	\rule{0pt}{3ex}    
	$
	%	\frac{\pi-2}{2 \pi} 
	\frac{1}{2} - \frac{1}{\pi}
	+ O(d^{-1/2})$ & $d -1$ \\
	\rule{0pt}{3ex}    
	$ \frac{1}{4} + O(d^{-1/2})$ & $d-1$ \\
	\rule{0pt}{3ex}    
	$
	%	 \frac{2 + \pi}{4 \pi} 
	\frac{1}{4} + \frac{1}{2\pi}
	+ O(d^{-1/2})$ & $\frac{d(d-3)}{2}$ \\
	\rule{0pt}{3ex}    
	$ \frac{d}{4} + \frac{1}{2} + O(d^{-1/2})$ & $d-1$ \\
	\rule{0pt}{3ex}    
	$ \frac{d}{4} + \frac{-4 + \pi + \pi^{2}}{2 \pi \left(-4 + \pi\right)} + 
	O(d^{-1/2})$ & 
	$1$ \\
	\rule{0pt}{3ex}    
	$ \frac{d}{\pi} + \frac{- 10 \pi + 8 + \pi^{2}}{2 \pi \left(-4 + 
		\pi\right)} + O(d^{-1/2})$ & $1$. \\
\end{tabular}	
\caption{The Hessian spectrum at global and spurious $\Delta (S_{d-p} \times S_p)$-minima of  $\ploss_{\text{ReLU}}$, $p\in \{0,1,2,3\}$, given to $O(d^{-\frac{1}{2}})$-terms  \cite{arjevani2021analytic}. Eigenvalues having large multiplicity  concentrate near zero and account for all but $\Theta(d)$ eigenvalues that grow linearly in $d$. Upon convergence, the spectral density is expected to accumulate in clusters whose number does not depend on $d$.}
\end{figure}

\subsection{Symmetric tensor norms}
In many cases of interest, for example, regularity conditions (see \propC~in \pref{sec: gen and const}) or nearly identical Hessian spectra (see \pref{table: skew}), second-order information may not suffice. Unfortunately, for phenomena involving $k$th order derivatives $(k\ge 3)$ of $C^k$ functions $f:\RR^d\to \RR$, the method described in \pref{sec: app rep} cannot be used directly as $d^k f_\c$ is no longer a $G$-map (relatedly, note that computing the spectral norm of high-order tensors is worst-case NP-hard \cite[10.2]{hillar2013most}). It is possible, however, to obtain bounds on the spectral norm of $D^k f(\c) \in \symp^k\RR^{d}$, the symmetric tensor representing $d^k f$ in the standard basis, via the operator norm of its unfoldings. To ease the exposition, we work in this section in coordinates (rather than with symmetric multilinear maps, see \pref{sec: local theory}) and assume the standard Euclidean inner product. We have, 
\begin{align}\label{tensor norm relax}
\|D^k f(\c)\|_\sigma &\le \sup\{\|D^kf(\c)\x^{\otimes (k-1)}\|_2~|~ \|\x\|_2\le1\} \nonumber\\
&\le  \sup \{ \|D^k f(\c) M \|_2~|~ M\in \symp^{k-1}(\RR^d), \|M\|_2\le 1\}.
\end{align}
If $f$ is $G$-invariant, then $D^k f(\c)$ is a $G$-map from $(\symp^{k-1}(\RR^d),\allowbreak G)$ to $(\RR^d, G)$ (the former defined in \pref{sec: reps}), and the method described in \pref{sec: app rep} may be used to compute \emph{singular} values of $D^k f(\c)$, or rather of an unfolding thereof (cf. \cite{hu2015relations}), granted all choices involved are orthonormal. For example, referring to $(\RR^d, S_d)$, the isotypic decomposition for  $k=3, 4$~are
\begin{align}
\symp^2(\RR^d) &= V_{(d)}^{\oplus 2} \oplus V^{\oplus 2}_{(d-1, 1)} \oplus V_{(d-2,2)},\label{iso_decomp_sym2}\\
\symp^3(\RR^d) &= V_{(d)}^{\oplus 3} \oplus V^{\oplus 4}_{(d-1, 1)} \oplus V^{\oplus 2}_{(d-2,2)} \oplus V_{(d-2, 1, 1)} \oplus V_{(d-3, 3)}, \label{iso_decomp_sym3}
\end{align}
holding, respectively, for $d\ge 4$ and $d\ge 6$. 
For $(M(d,d), S_d)$, we have,
\begin{align}\label{decomp_mdd}
M(d,d) \cong (\RR^{d})^{\otimes 2} &= V_{(d)}^{\oplus 2} \oplus V^{\oplus 3}_{(d-1, 1)} \oplus V_{(d-2, 1, 1)} \oplus V_{(d-2, 2)},\\
\symp^2((\RR^{d})^{\otimes 2})&= 
V_{(d)}^{\oplus 11}
\oplus V_{(d - 1 ,1 )}^{\oplus 21}
\oplus V_{(d - 2 ,2 )}^{\oplus 19}
\oplus V_{(d - 2 ,1 ,1 )}^{\oplus 13}
\oplus V_{(d - 3 ,3 )}^{\oplus 6}\nonumber\\
&\oplus V_{(d - 3 ,2 ,1 )}^{\oplus 10}
\oplus V_{(d - 3 ,1 ,1 ,1 )}^{\oplus 4}
\oplus V_{(d - 4 ,4 )}
\oplus V_{(d - 4 ,3 ,1 )}
\oplus V_{(d - 4 ,2 ,2 )}^{\oplus 2}\nonumber\\
&\oplus V_{(d - 4 ,2 ,1 ,1 )}
\oplus V_{(d - 4 ,1 ,1 ,1 ,1 )},	\\
\symp^3((\RR^{d})^{\otimes 2})&=
V_{(d)}^{\oplus 52}
\oplus V_{(d - 1 ,1 )}^{\oplus 147}
\oplus V_{(d - 2 ,2 )}^{\oplus 175}
\oplus V_{(d - 2 ,1 ,1 )}^{\oplus 157}\nonumber\\
&\oplus V_{(d - 3 ,3 )}^{\oplus 111}
\oplus V_{(d - 3 ,2 ,1 )}^{\oplus 183}
\oplus V_{(d - 3 ,1 ,1 ,1 )}^{\oplus 79}
\oplus V_{(d - 4 ,4 )}^{\oplus 36}\nonumber\\
&\oplus V_{(d - 4 ,3 ,1 )}^{\oplus 86}
\oplus V_{(d - 4 ,2 ,2 )}^{\oplus 54}
\oplus V_{(d - 4 ,2 ,1 ,1 )}^{\oplus 70}
\oplus V_{(d - 4 ,1 ,1 ,1 ,1 )}^{\oplus 20}\nonumber\\
&\oplus V_{(d - 5 ,5 )}^{\oplus 6}
\oplus V_{(d - 5 ,4 ,1 )}^{\oplus 16}
\oplus V_{(d - 5 ,3 ,2 )}^{\oplus 20}
\oplus V_{(d - 5 ,3 ,1 ,1 )}^{\oplus 18}\nonumber\\
&\oplus V_{(d - 5 ,2 ,2 ,1 )}^{\oplus 18}
\oplus V_{(d - 5 ,2 ,1 ,1 ,1 )}^{\oplus 12}
\oplus V_{(d - 5 ,1 ,1 ,1 ,1 ,1 )}^{\oplus 4}
\oplus V_{(d - 6 ,6 )}\nonumber\\
&\oplus V_{(d - 6 ,5 ,1 )}
\oplus V_{(d - 6 ,4 ,2 )}^{\oplus 2}
\oplus V_{(d - 6 ,4 ,1 ,1 )}
\oplus V_{(d - 6 ,3 ,3 )}^{\oplus 2}\nonumber\\	
&\oplus V_{(d - 6 ,3 ,2 ,1 )}^{\oplus 2}
\oplus V_{(d - 6 ,3 ,1 ,1 ,1 )}
\oplus V_{(d - 6 ,2 ,2 ,2 )}^{\oplus 2}
\oplus V_{(d - 6 ,2 ,2 ,1 ,1 )}^{\oplus 2}\nonumber\\
&\oplus V_{(d - 6 ,2 ,1 ,1 ,1 ,1 )}
\oplus V_{(d - 6 ,1 ,1 ,1 ,1 ,1 ,1 )},
\end{align}
holding, respectively, for $d\ge 4$, $d\ge 8$ and $d\ge 12$. In \pref{sec: stab chars}, we explain how such isotypic decompositions, with the particular feature of holding for all arbitrarily large values of $d$, may be obtained.

\subsection{Annihilation of minima and \suf sufficiency of jets}\label{sec: ann}
An outstanding question in deep learning (DL) concerns the ability of simple gradient-based methods to successfully train neural networks. Indeed, nonconvex optimization landscapes may have spurious (i.e., non-global local) minima with large basins of attraction and this can cause a complete failure of these methods. Evidence suggests that this problem can be circumvented by the use of a large number of parameters. The sharp analytic estimates of the Hessian spectrum given at SB critical points allow us to study how the loss landscape is transformed when the number of neurons is increased. For example, in two-layer ReLU networks \cite{arjevani2022annihilation}, $\Delta S_d$ and $\Delta (S_{d-1} \times S_1)$ spurious minima are shown to transform into saddles; the former by adding one neuron, the latter by two. In particular, annihilation of minima is seen to be tied to symmetry. (In both cases, the loss is essentially not affected by this process.)

Of course, a minimum of a function having  the same second-order Taylor polynomial as that of the model studied in op. cit. would undergo the same transition. Generalizing this, and in accordance with our general approach, our study of annihilation of minima proceeds by reducing to finite models determined by \emph{jets}, see \pref{sec: tan pre}. We say that a $k$-jet $\mu$ is \suf\emph{sufficient} if any $f$ satisfying $j^k_\c f =\mu$ have the same extremal type at $\c$ as $\mu$. Initial work aimed at characterizing \suf sufficiency include \cite{netto1984jet} which focused on real polynomials. We cover generalizations of the results as well as their genericity in the context of o-minimal theory in \cite[\secsuf]{arjevani2024sc2}. A stronger equivalence relation, $C^0$-sufficiency, holding for general $C^k$ functions requires the existence of a homeomorphism $h$ locally satisfying $T_0^k f = f\circ h$. $C^0$-sufficiency was original introduced by Kuo in \cite{kuo1969c0}, who later also defined the related, but generally not \suf-sufficient, $v$-sufficiency \cite{kuo1972characterizations} (the latter being implied by our use of ambient isotopy equivalence, \pref{sec:local_theory}). If the first nonvanishing homogeneous component satisfies certain regularity conditions, an elegant blowing-up construction employed by \cite{buchner1983applications} shows that $h$ may be selected as a $C^1$-diffeomorphism. Stronger still 
are \emph{contact} equivalences ubiquitous in singularity theory \cite{golubitsky2012stable} (and see \cite{ruas2022old} for a more recent account). However, these are generally too restrictive to establish genericity in the context of our work.

Once \suf-sufficiency of fixed degree has been established for a given class of functions, critical points may be classified by reducing to (finite) jets. In \secsuf, we carry out this program for the permutation representations considered in this work. Additional representations related to NNs, such as
convolutional layers and residual connections, as well as subgroups of $S_d$ given by the automorphism groups of the affine and the projective line over finite fields (see \cite[Example 3.4]{arjevanifield2022equivariant}) are studied in \cite{arjevani2025optsym} by use of methods from elimination and intersection theory (e.g., \cite{cartwright2013number} and toric geometry, see \pref{sec: poly tan}). For example, in \cite{arjevani2022annihilation} Hessian eigenvalues are computed by rewriting their Puiseux series in terms of the gradient entries (formally, quotienting out the ideal generated by the respective Puiseux terms). As gradient entries vanish at critical points, this allows for a direct evaluation of the eigenvalue expressions. The approach is used for studying other choices of kernels, in particular polynomial ones where the expressions of the gradient entries quotiented out in the respective ring are simpler, see \secsuf. We remark that additional algebraic relations between criticality, curvature and loss exist and indicate a certain rigidity of the loss landscape.

\subsection{Stabilizing inner products and families of critical points}
For kernels possessing certain geometric characteristics, including those described in (\ref{geo ker}), a path-based approach introduced in \cite{ArjevaniField2020} enables the construction of infinite families of critical points for (\ref{opt problem}), yielding one critical point for each sufficiently large $d\in\NN$. The approach generally assumes \emph{natural} sequences of groups $(G_d)$ and target matrices $(T_d)$ having the the property that: if $k-d$ is fixed, then inner products between rows of $W\in M(k, d)^{G_d}$ and $T_d$ are quadratic polynomials in $d$ and the $N$ variables given by the identification of $M(k, d)^{G_d}$ and $\RR^N$ through  a suitable linear isomorphism $\Xi\defeq \Xi(d):\RR^N\to M(k,d)^{G_d}$ (in particular the dimension of the latter stabilizes for $d$ sufficiently large). Thus, for natural sequences, if the kernel is given in terms of inner products of its arguments, then we may express $\ploss$ as a function of polynomials in $d$ and $\bxi\in\RR^N$, and so regard $d$ as a real variable. Assuming further that the kernel is \emph{definable} in an o-minimal structure (see  \pref{sec:omin_pre}), the set $\sigma_\ploss \defeq \{(\bxi, d)\in \RR^{N+1} ~|~\Xi(\bxi)\in \critset(\ploss(\cdot;d))\}$ is definable, hence so its image under the (Nash) map 
$\cV_d(\x) \defeq \prn*{{x_1}/{\sqrt{1+\|\x\|^2}},\dots,{x_d}/{\sqrt{1+\|\x\|^2}}}$ mapping $\RR^{d}$ isomorphically onto $(-1,1)^d$. Each infinite sequence of critical points $W_d = \Xi(\bxi_d)$, $d$ sufficiently large, gives a point in $\cl{\cV_{N+1}(\sigma_\ploss)} \cap \partial [-1,1]^{N+1}$ which, by the Curve Selection Lemma (CSL), may be approached by a definable arc lying in $\cV_{N+1}(\sigma_\ploss)$,  yielding in turn critical points in $M(k,d)$ (uniqueness may be established, generically, by standard transversality arguments). 

For globally subanalytic \cite{van1986generalization} kernels, the entries of the arc so constructed may be given by Puiseux series (in $1/d$). The series coefficients may then be computed in an obvious gradual manner (sharp, though not necessarily analytic,  estimates may also be obtained for the more general class of polynomially bounded o-minimal structures). By elementary results in o-minimal theory, the Hessian spectrum may also be given in terms of Puiseux series, see \pref{table: skew}. 

Finally, a consequence of $\sigma_{\ploss}$ being definable is that \emph{local triviality} applies and so the topological type of $\Xi^{-1}\critset(\ploss(\cdot;d))$ stabilizes for sufficiently large~$d$. For small values of $d$, curves of critical points may bifurcate, as they sometimes do.

\subsection{Deep symmetry breaking} \label{sec: app nns}
\begin{figure}[h]\label{fig: emp_evs}
\centering
\includegraphics[scale=0.25]{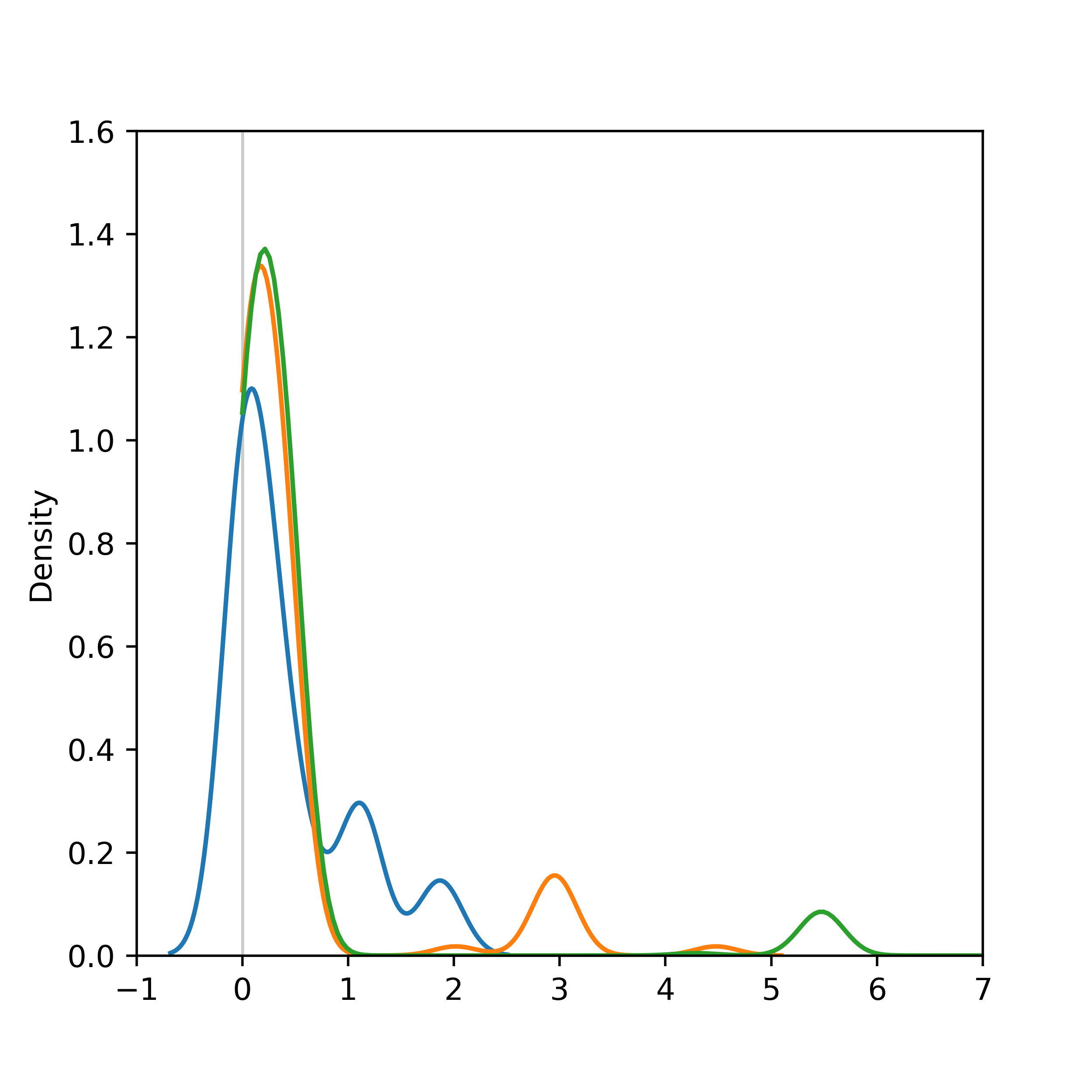}
\includegraphics[scale=0.45]{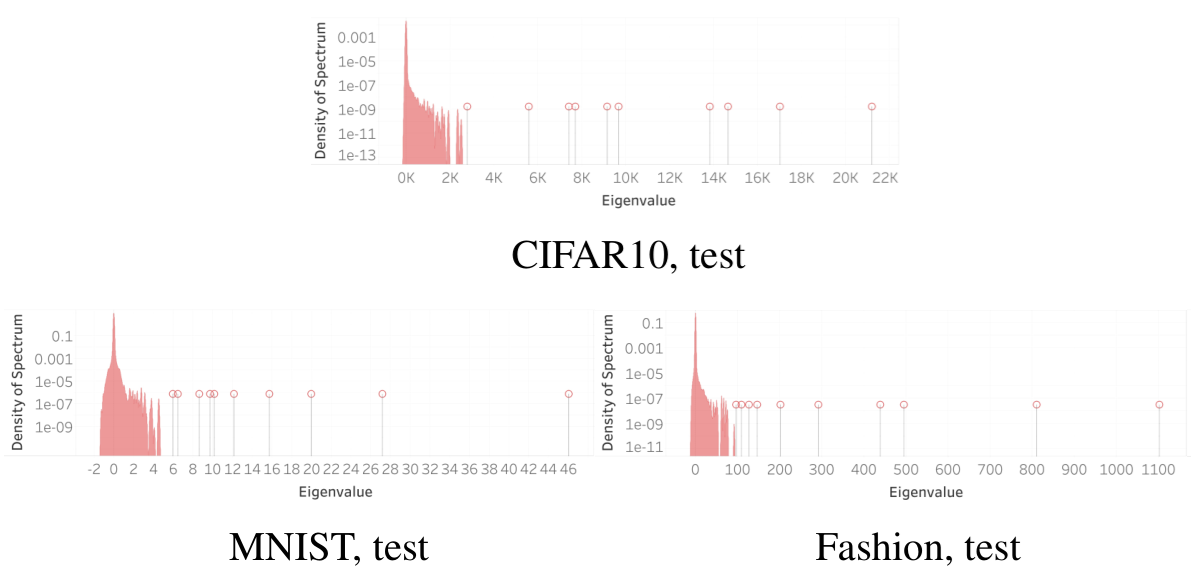}
\caption{(Left) an histogram representing the analytic estimates of the Hessian spectrum given in \pref{table: skew} \cite{arjevani2021analytic}. By fundamental results from group representation theory, symmetry breaking critical points necessarily have highly skewed Hessian spectrum. (Right) numerical approximation of the Hessian spectrum for VGG11 trained on various datasets \cite{papyan2018full}. In \cite{arjevani2024dsb}, the mechanism developed in this work is used to examine the extent to which SB phenomena account for the highly skewed Hessian spectrum observed in practice.}
\end{figure}

For a large class of group representations, for example, those considered in \pref{sec: stab chars}, the Hessian spectrum is highly skewed at SB critical points on account of the small number of terms occurring in the respective isotypic decomposition, see \pref{table: skew}. Qualitatively, the same phenomenon has been observed across a variety of empirical studies involving different network architectures and datasets, e.g., \cite{sagun2016eigenvalues,chaudhari2019entropy}, and see \pref{fig: emp_evs}. The extent to which the latter indicates the presence of SB critical points is studied in \cite{arjevani2024dsb} for various realistic settings based on the approach developed in this work. Some of the symmetries involved in the analysis are naturally continuous (Lie). These are addressed in \cite{arjevani2025optsym}, along with aspects related to~dynamics.

% !TEX root = general_symmetry_and_critical_points.tex

\section{The tangency set}

\subsection{Preliminaries} \label{sec: tan pre}
We briefly review concepts from differential topology and o-minimal theory, and fix notation that is used throughout.

In the intended applications, insofar as the local study of tangency sets is concerned, functions are often, but not always, locally smooth ($C^\infty$). In some cases, the differentiability order may be finite (for example, global minima of $\ploss_{\mathrm{ReLU}}$), and the results hold with the appropriate restrictions.

A second category of the results pertains to the global aspects of the tangency set, e.g., stratification and local triviality, construction of arcs giving infinite families of critical points, and connectedness, see 
\cite[\secsuf~and \seccon]{arjevani2024sc2}. Here, even finite differentiability may be too strong of an assumption. Therefore, to maintain a reasonable structure of the tangency set, we compensate for a possible lack of differentiability by requiring definability in an o-minimal structure, see \pref{sec:omin_pre}.

\subsubsection{Transversality theory}\label{sec: local theory}
For introductory texts covering the topics in the sequel, we refer the reader to  \cite{gibson2006topological,hirsch2012differential,golubitsky2012stable}. We usually do not indicate the order of a manifold so as to avoid the tedious bookkeeping. However, these will always be consistent.

Let $M$ and $N$ be manifolds and $1\le r\le\infty$. We denote the $\RR$-algebra of $C^r$ real-valued functions on $M$ by $C^r(M)$, and  the set of $C^r$ maps from $M$ to $N$ by $C^r(M, N)$ (the latter being a $C^r(M)$-module if $N$ is a vector space). Given a point $x\in M$, a \emph{germ} at $(M,x)$ is an equivalence class of $C^r(M, N)$-maps defined in neighborhoods of~$x$, two maps being equivalent if they agree in some neighborhood of~$x$. Germs are similarly defined for sets, e.g., tangency sets. A map germ at $x$ mapping the point $x$ to the point $y$ in $N$ is denoted by $f:(M,x)\to (N,y)$. We use maps from a given germ to refer to the germ itself. For the concepts involved in our analysis, this ambiguity shall cause no confusion.

If $f\in C^r(M, N)$ and $r <\infty$, we denote the $r$-jet of $f$ at $x\in M$ by $j^rf(x)$ or $j_x^rf$. We call $x$ the source and $f(x)$ the target of the jet. The bundle of $r$-jets of $C^r(M, N)$-maps  is denoted by $J^r(M,N)$; if the source $x$ is fixed, we write $J^r_x(M,N) $. We denote by $L_s^k(\RR^m, \RR^n)$ the vector space of $k$-multilinear symmetric maps (forms) from $\RR^m$ to $\RR^n$, and $P^{k}(\RR^m, \RR^n)$  (reps. $P^{(k)}(\RR^m, \RR^n)$) the space of $k$-degree (resp. $k$-degree homogeneous) polynomial maps  from $\RR^m$ to $\RR^n$. We write $P^{(k)}(\RR^m)\defoo P^{(k)}(\RR^m,\RR)$ and let $P(\RR^m) \defoo \cup_{k\ge 0} P^{(k)}(\RR^m)$ denote the (graded) $\RR$-algebra of real-valued polynomial functions on $\RR^m$. 
Recall that $L_s^k(\RR^m, \RR^n) \approx  P^{(k)}(\RR^m,\RR^n) $ by the natural \emph{restitution} isomorphism $A\mapsto A(\x^k)/k!,~A\in L_s^k(\RR^m,\RR^n)$,  $\x^k\defoo  (\x,\dots,\x)$, $\x$ repeated $k$ times. The reverse direction of this isomorphism is called \emph{polarization}. Given a $k$-degree homogeneous polynomial $p\in P^{(k)}(\RR^m)$, the \emph{polar form} of $p$ is the  $k$-multilinear symmetric form $A$ defined by 
\begin{align*}
A(\x_1,\dots, \x_k) = \frac{1}{k!} \frac{\partial}{\partial \lambda_1}
\cdots \frac{\partial}{\partial \lambda_k} p(\lambda_1 \x_1 + \cdots + \lambda_1\x_k)|_{\lambda_1 = \cdots =\lambda_k=0}.
\end{align*}
Thus $P^{r}(\RR^m, \RR^n) \approx \RR^n \times  L(\RR^m, \RR^n)\times L_s^2(\RR^m, \RR^n)\times \cdots \times L_s^r(\RR^m, \RR^n)$. In the special case $M=\RR^m$, $N=\RR^n$, if $U\subset \RR^m$ is open and $f\in C^r(U,\RR^n)$, the $r$-jet of $f$ at $x$ has a canonical representative in $P^{r}(\RR^m, \RR^n)$, namely, its $r$-order Taylor polynomial map $T^r_xf$ at $x$. Thus,
\begin{align}\label{jet_identification}
J_x^r(\RR^m, \RR^n) &=  P^{r}(\RR^m, \RR^n),\\
J^r(\RR^m, \RR^n) &= \RR^m\times P^{r}(\RR^m, \RR^n).\nonumber 
\end{align}
Under this identification, if $f\in C^r(\RR^m,\RR^n)$, then 
\begin{align*}
j^rf(x) = (x, f(x), Df(x), \dots, D^rf(x)),
\end{align*}
where $d^kf_x \in L_s^k(\RR^m, \RR^n)$ is the $k$th derivative of $f$ at $x$. 

We use the $r$-jet \emph{extension} map $j^r:C^r(M,N)\to C^0(M, J^r(M,N))$, mapping $f$ to $j^r f$, to define the $C^r$- and the (stronger) Whitney $C^r$-topologies as ones induced by $j^r f$: the former with respect to the compact-open topology on $C^0(M, J^r(M,N))$, the latter with respect to the topology generated by basic open sets of the form $\cN(f,U)\defoo \{g\in C^0(M, J^r(M,N)) ~|~ \Gamma_g \subseteq U\}$, where $U\subseteq M\times J^r(M,N)$ is an open set containing $\Gamma_f$, the graph of $f$. In the $C^\infty$ category, analogous definitions, both topologies, follow by regarding $J^\infty(M,N)$ as the inverse limit of finite jet spaces. If $M$ is compact, $1\le r\le \infty$, both topologies coincide. If $M$ is not compact, Whitney $C^r$ topology is extremely large. Nevertheless, the category theorem of Baire is valid in both topologies (the former being a complete metric space by Baire category theorem, the latter by, e.g., \cite{mather1970stability}). 

\subsubsection{Transversality and stratification} If $f:M\to N$ is a $C^r$ map ($r\ge 2$) and $P\subseteq N$ a submanifold, we write $f \trans P$ to mean that $f$ is \emph{transverse} to $P$. That is, if $x\in M$ and $f(x)\in P$, then $df_x(T_xM) + T_{f(x)}P = T_{f(x)}N$. It is well-known that if $f\trans P$ then $f^{-1}(P)\subseteq M$ is a manifold and $\codim(f^{-1}(P)) = \codim(P)$. We say that $f$ is transverse to $P$ \emph{along} $A\subseteq M$ to indicate that transversality is only required at $x\in A\cap f^{-1}(P)$. 
\begin{definition}
A \emph{stratification} $\fX$ of a subset $X\subseteq M$, $M$ a manifold, is a locally finite partition of $X$ into submanifolds, called \emph{strata}. We say that the pair $(X,\fX)$ is a \emph{stratified set}.
\end{definition}
If $Y\subseteq N$ and $(Y,\fY)$  is a \emph{stratified} set, one requires stratumwise transversality: $f\trans Y$  if $f\trans A$ for all $A\in\fY$. Thus if $f\trans Y$, $\fY$ pulls back to a stratification $f^*\fY$ \emph{induced} on~$f^{-1}(Y)$. 

Of importance are \emph{Whitney regular} stratifications yielding local triviality. As the notions involved in the definition are diffeomorphism invariants we may invoke Whitney embedding theorem and so assume that $M$ and $N$ are closed embedded submanifolds of $\RR^m$ and $\RR^n$, respectively. The set of $k$-dimensional vector subspaces of $\RR^n$ naturally embeds (Pl\"{u}cker embedding) into $\PP\prn{\bigwedge^k \RR^n} $, the projectivization of the $k$th exterior product of $\RR^n$, as a nonsingular projective algebraic variety. The image of Pl\"{u}cker embedding together with the subset topology (compact) is called the Grassmannian $\bG(k, \RR^n)$. For $k=1$, one recovers the projective space $\bG(1, \RR^n) = \bP(\RR^n)$. We note that convergence in $\bG(k, \RR^n)$ can be equivalently defined in terms of distances between the orthogonal transformations that project onto the $k$-dimensional~subspaces. 

\begin{definition}
Let $M$ be a manifold and $X\subseteq M$. 
\begin{itemize}[leftmargin=*]
\item A stratification $\fX$ of $X$ is 
Whitney (a)-regular if for any two strata $A,B\in\fX$ and a sequence $(b_k)\subseteq B$ converging to $a\in A$ such that $(T_{b_k}B) \to T$ in $\bG(\dim(B), \RR^n)$,  $T_a A\subseteq T$.
\item A stratification $\fX$ of $X$ is Whitney regular (or (b)-regular) if for any two strata $A,B\in\fX$  and sequences $(a_k)\subseteq A$,  $(b_k)\subseteq B$ with $a_k\to a$ and $b_k\to a$, the following holds: if $(T_{b_k}B) \to T$ in $\bG(\dim(B), \RR^n)$ and $\RR(b_k -a_k) \to l$ in $\PP(\RR^n)$, then $l\subseteq T$.
\end{itemize}
\end{definition}

\begin{rem} \label{rem:frontier}
Whitney regularity implies Whitney (a)-regularity, which in turn characterizes stability \cite{trotman1978stability}. If $X$ is closed and the strata are connected, the former further implies the frontier condition, namely, $A\cap \closure{B} \neq \emptyset\implies A\subseteq \closure{B}$ \cite[Corollary 10.5]{mather2012notes}. Moreover, for such $A$ and $B$, $\dim A < \dim B$ \cite[Proposition 2.7]{mather2012notes}.
\end{rem}

\begin{prop}(\cite[Corollary 8.8]{mather1973stratifications}) If $\fY$ is a Whitney regular stratification of $Y\subset N$  and $f\trans Y$, then $f^*\fY$ is Whitney regular.
\end{prop}

\begin{thm}\label{thm:thom_mather}[Thom-Mather Transversality] Let $M,N$ be smooth manifolds and $Y$ a closed Whitney stratified subset of $N$.
\begin{enumerate}[leftmargin=*]
\item If $f\in C^r(M,N)$ and $f\trans Y$ at $x\in M$, then $f\trans Y$ on some neighborhood of $x$. 

\item If $f\in C^r(M,N)$, then $\{x\in M~|~f\trans Y \text{at } x\}$ is an open subset of~$M$.
\item The set $\{f\in C^r(M, N)~|~f\trans Y\}$ is residual (hence dense) in the Whitney-$C^r$ topology.
\item If $M$ is compact and $f:M\times [0,1]\to N$ is a family of maps such that $f_t\trans Y, t\in [0,1]$, then there exists a $C^0$-isotopy $F:M\times [0,1]\to M$ of homeomorphisms of $M$ such that $F_t(f_t^{-1}(Y)) = f_0^{-1}(Y), t\in [0,1]$.
\item (Jet transversality) Statements analogous to (1-4) hold for $f\in C^r(M,N)$ satisfying $j^r f\trans Y$, where $Y\subseteq J^r(M,N)$ is a closed Whitney stratified set.
\end{enumerate}
\end{thm}

The set of Whitney stratifications may be endowed with a partial order.
\begin{definition}[Notation and assumptions as above]
Associate to a Whitney regular stratification $\fX$ a \emph{filtration by dimension}  $X=\cup_i X_i$, where $X_i$ denotes the union of strata of dimension $\le i$. If $\fX\neq \fX'$ are Whitney stratifications and $(X_i), (X_i')$ the respective associated filtrations by dimension, 
we say that $\fX< \fX'$ if there exists $i$ such that $X_i\subsetneq X_i'$ and $X_j = X_j'$ for any $j> i$.
\end{definition}
We refer to a minimal stratification (unique) as \emph{canonical}. Although a minimal stratification may not always exist, it does for important classes of sets \cite[Chapter 4]{molina2020handbook}, notably, semialgebraic and semianalytic sets.
\begin{thm}
Every semialgebraic (resp. semianalytic) subset $X$ has a minimal Whitney stratification $\fX$ with finitely many semialgebraic (resp. semianalytic) strata  \cite{mather1973stratifications}.
\end{thm}

\subsubsection{O-minimal theory}\label{sec:omin}
A far-reaching generalization of semialgebraic and subanalytic geometry is given by sets {definable} in a \emph{o-minimal structure} \cite{van1996geometric}. Like their semialgebraic counterparts, definable sets Whitney (indeed, Verider) stratify definably. This leads in particular to a well-behaved dimension theory (see also the fundamental \emph{cylindrical definable cell} decomposition). 

\begin{definition} \label{def:omin}
	An o-minimal structure on $(\RR, +, \cdot)$ is a sequence $\cD = (\cD_n)_{n\in\NN}$ such that for each $n\in\NN$:
	\begin{enumerate}
		\item[(D1)] $\cD_n$ is a boolean algebra of subsets of $\RR^n$, i.e., $\cD_n$ is closed under taking complements and finite unions.
		\item[(D2)] If $A\in\cD_n$, then $A\times \RR$ and $\RR\times A$ are in $\cD_{n+1}$.
		\item[(D3)] If $A\in\cD_{n+1}$ then the projection on the first $n$ coordinates $\pi(A)$ is in $\cD_n$.
		
		\item[(D4)] $\cD_n$ contains $\{\x\in\RR^n~|~P(\x)=0\}$ for every polynomial $P\in \RR[X_1,\dots,X_n]$.
		\item[(D5)] \emph{(o-minimality)} Each set  belonging to $\cD_1$ is a finite union of intervals and points.
	\end{enumerate}
\end{definition}
Notable examples of o-minimal structures are given by  semialgebraic sets and globally subanalytic sets. The latter can be equivalently defined as follows. Denote by $\RR_{\mathrm{an}}^K$ the o-minimal structure generated by addition, multiplication, analytic functions restricted to $[-1,1]^d$ and the power functions $x\mapsto x^r$ on $(0,\infty)$ for some $r\in K$, $K\subseteq \RR$ a fixed field. Then, the o-minimal structure of globally subanalytic sets is precisely $\RR_{\mathrm{an}}\defoo \RR_{\mathrm{an}}^\QQ$. Additional o-minimal structures may be obtained, for instance, by taking $\RR_{\mathrm{an}}^\RR$ or, by a well-known result, allowing the exponential function $\RR_{\mathrm{an}, \exp}^K$ (and, more generally, Pfaffian functions).

A set $X\in\RR^n$ is \emph{definable} if $X\in\cD_n$. A map $F:X\arr\RR^n$ is called \emph{definable} if its graph  $\Gamma_F$ is definable. Sets defined by first-order formulae ranging over {definable sets} are themselves definable. In particular, constructions such as the inverse of a definable function or the sum, product or composition of definable functions, being expressible  in terms of first-order formulae, preserve definability. Thus, for example, if $f:\RR^d\arr\RR$ is definable (not necessarily $C^1$), the set 
\begin{align}\label{minimizers}
\minimizers \defoo \cup_{r>0} \argmin f|_{\sphere_\c(r)}
\end{align}
consisting of all points where $m_\c(r)\defoo \min\{f(\x)~|~\x\in\sphere_\c(r)\}$ is attained, can be  given by 
\begin{align}
	\label{eqn:logic_minimizers}
	\{\x\in\RR^d~|~\exists r\in \RR^{>0}, \forall \y\in\RR^d &(\|\y-\x\|^2 = r^2 \\
	&\implies \exists s\in\RR, f(\y) - f(\x) = s^2)   \}\nonumber
\end{align}
and so is definable (leaving the validation of use of abbreviations such as $\RR^{>0}$ and $f(\x)-f(\y) = s^2$ to the reader). Likewise, $M_\c(r)$ giving the maximum of $f$ over spheres centered at $\c$ is definable. 
Throughout, all sets and mappings involved in the analysis are definable when $f$ is. The proofs are straightforward and so are generally omitted. 

We describe several fundamental results in o-minimal theory that are used in later sections.

\begin{prop}[Monotonicity theorem] \label{prop:monotonicity}
	\textit{Let $f:(a,b)\to\RR$ be a definable function, $-\infty\le a<b\le \infty$, and $k\in\NN$ fixed. There are $a_0,\dots,a_{m+1}$ with $a=a_0<a_1<\cdots<a_{m+1} = b$ such that $f|_{(a_i,a_{i+1})}$ is $C^k$, and either constant or strictly monotone for $i=0,\dots,m$.}
\end{prop}

\begin{prop}[Curve Selection Lemma (CSL)]
If $x\in \cl{X}$, $X$ definable, then there exists a definable continuous map $\gamma:[0,1)\to X$ such that 
$\gamma(0)= x$, $\gamma(0,1)\subseteq X$ and $\nrm{\gamma(s) - x} = s$. 
\end{prop}
By the Monotonicity theorem, we can always choose $\gamma$ to be  $C^k$. If $X$ is definable in $\RR_{\mathrm{an}}$ then $\gamma$ may be extended to an analytic function on $(-1,1)$ (a fact used later when polynomial tangency sets are considered), but this fail in general for $\RR_{\mathrm{an}}^\RR$ and $\RR_{\mathrm{an},\exp}$. 

An o-minimal structure $\cD$ is \emph{polynomially bounded} if for every $\cD$-definable function $f:\RR\to\RR$,  $f(t)=O(t^n)$ for some $n\in \NN$ as $t\to \infty$. 
\begin{prop}[Growth Dichotomy]
Either $\cD$ is polynomially bounded or $\cD$ contains the exponential function. In the former case, either $f(t) = ct^r + o(t^r)$ for some $c,r\in\RR$ as $t\to0^+$, where $x\mapsto x^r$ is definable in $\cD$, or $f$ ultimately vanishes identically.
\end{prop}

Fundamental results in calculus, for instance, the implicit function theorem, often have definable versions that maintain the definable data. In some cases, like Morse-Sard's theorem \cite{loi2010lecture}, some of the hypotheses can be omitted. This equally applies to fundamental results in semialgebraic geometry.

\begin{definition}
Let $E,B$ be definable sets. A definable map $f:E\to B$ is \emph{definably trivial} if there are a definable set $F$ and a definable homeomorphism $h:E\to B\times F$ such that $f(x) = \pi_B\circ h(x), x\in E$, where $\pi_B$ is the obvious projection onto the first factor. 
\end{definition}

\begin{prop}[Local triviality]
Given a continuous definable map $f:E\arr B$ between definable sets we may partition $B$ into finitely many definable sets $B_1,\dots,B_m$ such that $f|_{f^{-1}(B_i)}, i\in [m],$ are definably trivial.
\end{prop}

\subsection{Regularity and jet transversality}\label{sec:local_theory}
The concept of tangency set naturally arises in the study of singularities (under different names), for examples  \cite{netto1984jet,nemethi1992milnor,le1998bifurcation,durfee1998index,pham2016genericity}. Our use is different in that we focus on properties of critical points connected by the tangency set, their symmetry and stability in particular, rather than, e.g., phenomena occurring at infinity such as failures of local trivialization. 

Our approach to studying germs of tangency sets involves formulating the problem in terms of transversality conditions on the jet extension map. As a first natural attempt, observe that the tangency set can be equivalently defined as $j^1f^{-1}(Z_\c)$, where
\begin{align}\label{def:masterset}
\masterset_\c \defoo \{(\x, \g) \in \RR^d\times \RR^d~|~\rk\brk{\x-\c~|~\g} \le 1 \}
\end{align}
is regarded as a subset of $J^1(\RR^d, \RR)$ by identifying the pairs $(\x,\g)$ with $(\x, 0, \g)$
(of course there is no harm in assuming $f(\c) = 0$ when studying  tangency sets).
Modulo the choice of the center $\c$, $\masterset_\c$ is a \emph{determinantal} variety (affine or projective) containing all $2\times d$ matrices of rank at most 1. The variety carries a natural action of $GL(2)\times GL(d)$, preserving rank. Thus, by known results, a Whitney stratification of $\masterset_\c$ (minimal in this case) may be given by the orbit types of the action: $\masterset_\c \backslash\{0\}$ and $\{0\}$ (the singular locus). Requiring $j^1f \trans \masterset_\c$ yields openness, density and isotopy results for the class of function considered. However, as $\codim \{0\} = 2d+1$, this rules out functions having a critical point at the origin. One possibility to address this issue is to require instead that $j^1f \trans \masterset_\c\backslash\{0\}$, but lose openness (density still holds). Of course, formulating the tangency set as a transversal intersection with $\masterset_\c$ fails to capture the behavior at critical points as criticality at a fixed $\c$ is not a property stable under perturbations. Therefore, we restrict to the ideal of functions $\fspacec{}{U}{\c} \subseteq C^\infty(\RR^d)$, $U\subseteq \RR^d$ an open set containing $\c$, of smooth functions $f$ with $f(\c)$ and $df(\c)$ vanishing. Tangency sets taken relative to other points or functions with finite differentiability order are handled similarly. 

\begin{rem}
It is also possible to stratify $\masterset_\c$ so that $(0,0,0)\in\RR^d\times\RR\times \RR^d$ lies on a stratum of codimension $d$ by defining $\masterstrat_\c = \{\masterstrat_\c^{d+1}, \masterstrat_\c^{d+2}\}$, with $\masterstrat_\c^{d+2} \defoo \masterstrat_\c \cap (\RR^d\backslash\{0\}) \times \RR\times \RR^d$ a $(d+2)$-dimensional manifold and $\masterstrat_\c^{d+1} \defoo  0\times \RR\times \RR^d$ $(d+1)$-dimensional. However, $\masterstrat_\c$ is not Whitney regular. To see this, observe that 

\begin{align*}
T_{(\x,y,\g)} \masterstrat_\c^{d+1} &= 0\times \RR\times \RR^d.\\
T_{(\x,y,\g)} \masterstrat_\c^{d+2} &= \{(\u,y, \frac{g_{\mathrm{nz}}}{x_{\mathrm{nz}}}\u+\eta\x)~|~\u\in\RR^d,y,\eta\in\RR\},
\end{align*}
where $\mathrm{nz}$ denotes an index of a non-zero entry of $\x$. Take, for instance, $\tau = (0,0,\e_1)\in T_{(0,0)}\masterstrat_\c^{d + 1}$. Although $b_n = (\e_2, 0, \e_2)/n$ converges to $(0,0,0)\in \overline{\masterstrat_\c^{d+2}}\cap \masterstrat_\c^{d+1}$, $\tau$ does not lie in the $(d+2)$-dimensional space to which $T_{b_n}\masterstrat_\c^{d+2}$ (constant) converges. Therefore, $\masterstrat_\c$ is not even (a)-regular. By \pref{rem:frontier}, another way to see that $\masterstrat_\c$ is not Whitney (a)-regular is by observing that $j^1\|\cdot\|^2\trans \masterstrat_\c^{d+1}$ at $0$ but not to $\masterstrat_\c^{d+2}$ at any neighborhood of $0$. 
\end{rem}

By standard results (e.g., \cite[p. 358]{golubitsky1978introduction}), any $f\in \fspacec{}{U}{\c}$ may be given by $f(\x) = \sum_{i=1}^d h_i(\x)(x_i-c_i)^2$ with $H = (h_1,\dots,h_d)\in C^\infty(\RR^d, \RR^d)$.  Thus, we may define $u:J^1(\RR^d,\RR^d)\to J^1(\RR^d,\RR)$ by $u(\x, h^0, H^1) = (\x, \sum_{i=1}^d h^0_i (x_i-c_i)^2,  \brk{2h^0_i (x_i-c_i) + \sum_{j=1}^d H_{ji}^1 (x_j-c_j)^2}_{i=1}^d)$ and so have $j^1 f$ decomposing as $j^1 f = u \circ j^1 H$. In particular, $\tanset_\c(f) = j^1 f^{-1}(\masterset_\c) = (j^1 H)^{-1} (u^{-1} (\masterset_\c))$. 
\begin{lemma}
The set $u^{-1} (\masterset_c)$ is a closed algebraic set of codimension~$d-1$.
\end{lemma}
\proof Trivial. \qed

\begin{definition}[Regularity, notation as above] \label{def: reg}
Let $\uniset_e$ denote the canonical stratification of $u^{-1}(\masterset_\c)$, the \emph{universal} stratification. If $j^1 H\trans \uniset_e$, we say that $\tanset_{\c}(f)$ is regular and that $f$ is $\tanset_{\c}$-regular. A regularity of the $\tanset_{\c}(f)$-germ at $\c$ is likewise defined using the function germ. In both cases, if $\c=0$, we usually simply write $\tanset$ rather than $\tanset_\c$.
\end{definition}

\begin{rem}
In the symmetric case, regularity is defined similarly, now accounting for the general form of invariant functions \cite{schwarz1975smooth}.
\end{rem}

For most of the development of the local theory, it is no loss of generality to assume that tangency sets are taken relative to the origin and focus on $\fspace{}{U}$, $U$ an open set containing the origin, and   $\prescript{}{d}{\gspace{e}}$, the ring of $\fspace{}{U}$-germs rooted at the origin. When results hold independently of $d$, we usually write $\gspace{e}$.  
\begin{cor}\label{cor:genericity} [Genericity]
The set of $\tanset$-regular functions (resp. germs) is dense and open in $\fspace{}{U}$ (resp. $\gspace{e}$), Whitney topology. 
\end{cor}
\proof By \pref{thm:thom_mather}, the set of $H$-maps giving regular tangency set is dense and open in $C^\infty(\RR^d,\RR^d)$. The map taking $H = (h_1,\dots, h_d) \to \sum_{i=1}^d h_i(x)x_i^2$ in $\fspace{}{U}$ is a continuous ($C^\infty$ topology) linear surjection of Fr\'echet spaces, hence open by the Open Mapping Theorem. Density and opennes in $\fspace{}{U}$ and $\gspace{e}$ now follow. \qed
\begin{cor}[Stability] \label{cor:ns_stab} 
If a $\tanset$-regular $f\in \gspace{e}$ is deformed into an $\tilde{f}\in \gspace{e}$ sufficiently close to $f$, then $\tanset(f)$ and $\tanset(\tilde{f})$ are isotopic.
\end{cor}
\proof By \pref{cor:genericity}, there exists a neighborhood $U$ of $f$ such that any continuous deformation of $f$ lying in $U$ describes a path of $\tanset$-regular functions. By  \pref{thm:thom_mather}, the associated tangency sets are isotopic. \qed

\begin{prop}\label{cor:tanset_is_one_dim}
If $\tanset(f)$ is regular and $\uniset_e'$ denotes the stratum containing $j^1H(0)$, then $\uniset_e'$ is of codimension $d-1$ or $d$. In the former case, $\tanset(f)$ is a one dimensional manifold. In the latter, a union of finitely many connected one-dimensional manifolds with the origin being a common boundary.
\end{prop}
\proof The first case is a simple consequence of regularity and the top stratum being of codimension $d-1$. For the second case, observe that points giving a minimum or a maximum of $f$ over spheres of sufficiently small radius always lie $\tanset(f)$ and so $0$ is never isolated. Thus, $j^1H(0)$ must lie in the closure of a top stratum. Since $\uniset_e$, being canonical, has finitely many strata, pulling back to the induced stratification $j^1H^{-1}(\uniset_e)$, Whitney regular, the claim follows.  \qed
\begin{exam} \label{exam:deg}
If $f(x_1,x_2) = \frac{1}{2}(x_1^2 +x_2^2)$ then $\tanset_0(f) = \RR^2$, and so by \pref{cor:tanset_is_one_dim} $f$ is not $\tanset_0$-regular. This can also be seen by noticing that $\tanset_0(f)$ is not stable in the sense of \pref{cor:ns_stab}.
\end{exam}

\subsubsection{Finite determinacy}

More can be said about regular tangency sets than what is stated in \pref{cor:tanset_is_one_dim}. However, in many cases, especially when symmetry is involved, obtaining a detailed description of the universal stratification $\uniset_e$ (see \pref{def: reg}) can become quite a formidable task. Thus, rather, our study of tangency sets proceeds by reducing to finite models given by Taylor polynomials. Of course, some caution is needed with respect to the degree of truncation. 

\begin{exam} \label{exam:taylor_not_regular}
Let $g_d:\RR^d\to\RR$ be a polynomial function defined by
\begin{align}
g_d(\x) = \frac{1}{2}\sum_{i=1}^d x_i^2+ \frac{1}{3}\sum_{i=1}^d x_i^3.
\end{align}
Computing, we find that $\tanset_{0}(g_d) = \{ \x\in\RR^d~|~ \x \in  \{0,c\}^d,~ c\in\RR\}$. However, referring to \pref{exam:deg}, since $T_0^2g_d(\x) = \frac{1}{2}\|\x\|^2$, $\tanset_{0}(T_0^2g_d) = \RR^d$, obviously not isotopic to $\tanset_{0}(g_d)$.
\end{exam} 

The situation described in \pref{exam:taylor_not_regular} becomes even more pronounced for \emph{flat} functions (derivatives of all order exist and vanish), such as $(x,y) \mapsto e^{-1/x^2} + e^{-1/y^2}$ (vanishing at the origin). This naturally leads to the notion of \emph{finite determinacy} building on fundamental concepts  in singularity theory, see e.g.,
\cite{golubitsky2012singularitiesI} (and \cite{mather1968stability} for initial work).

\begin{definition}[Finite determinacy]
Let $f\in\gspace{e}$ be $\tanset$-regular. If $T^r_0 f$ is $\tanset$-regular then we say that the set germ $\tanset(f)$ is $r$-determined ($r$ is not required to be minimally so).
\end{definition}
\begin{cor}\label{cor:finite_det}
If $\tanset(f)$ is $r$-determined then  $\tanset(f)$ and $\tanset(T_0^rf)$ are isotopic.
\end{cor}
\proof Define $F:U\times[0,1]\to \RR$ by $$F(\x, t) = tT^r_0 f(\x) + (1-t)f(\x),$$ with $U\subseteq \RR^d$  an open set where both $f$ and $T^r_0 f$ are defined, and invoke \pref{thm:thom_mather}.
\qed
\begin{rems}
1) The zero set of $F$ is sometimes referred to as the \emph{deformation variety}. The variety frequently arises in the context of finite determinacy, see e.g. \cite{kuo1972characterizations}.\\
2) By a well known results, algebraic sets in $\RR^d$ defined by polynomial equations of bounded degree belong to a finite list of homeomorphism types. By isotopy, so do finitely $r$-determined tangency~sets, $r$ bounded.\\
3) For definable functions, equivalence to jets may be established by Lojasiewicz's inequality.
\end{rems}

\subsection{Polynomial tangency sets and the curve selection lemma} \label{sec: poly tan}

We denote by $P(\RR^d)_0$ the subspace of polynomial vanishing and having a critical point at the origin.  If $p\in P(V)_0$ is $\tanset$-regular, we may use the CSL to construct analytic tangency arcs $\gamma$  locally parameterizing the one-dimensional manifolds given by  \pref{cor:tanset_is_one_dim}. The defining equation is 
\begin{align} \label{tan_poly}
d \prn*{p -\frac{\eta(t)}{2} \|\cdot\|^2}_{\gamma(t)} = 0,
\end{align}
where $\eta(t)$ is real analytic. We use (\ref{tan_poly}) to analyze different aspects of $\gamma$, hence of the $\tanset(p)$ germ. We may write 
\begin{align} \label{hom_comps}
p(\x) = \sum_{k=2}^{\deg(p)} A_k (\x^k),
\end{align}
where $\deg(p)\ge2$ and $A_k \in L_s^k(\RR^d, \RR)$ are the polarization of the respective $k$-homogeneous parts. Thus, $dp_\x(\cdot) = \sum_{k=2}^{\deg(p)} kA_k(\x^{k-1},\cdot)$. 
To simplify the notation, we shall identify $A_k(\x^{k-1},\cdot)$ with the \emph{column} vector associated by the standard Euclidean inner product  and remove parenthesis when no confusion results. 
\begin{definition}[Tangency equations]\label{def: tan_eqns}
Suppose $\gamma(t) = \sum_{i=1}^{\infty} t^i\v_i$ is normalized such that $\|\v_1\| =1$ and let $\eta(t) = \sum_{i=0}^{\infty} \eta^{(i)}t^i$. Equating like coefficients in (\ref{tan_poly}), the $k$th-order \emph{tangency equation} is given by the term corresponding to $t^k$.
\end{definition}

For example, the $k$th tangency equation for $k=1,2,3,4$ is, respectively,
\begin{align}
&2A_2\v_1 - \eta^{(0)} \v_1 = 0,\label{tangency_eqns}\\
&2A_2\v_2 +  3A_3\v_1^2 - \eta^{(0)} \v_2  - \eta^{(1)}\v_1   = 0,\nonumber\\
&2A_2\v_3   + 6 A_3(\v_1, \v_2) + 4 A_4\v_1^3 - \eta^{(0)} \v_3
- \eta^{(1)} \v_2 - \eta^{(2)} \v_1 = 0. \nonumber\\
&2 A_{2} \v_{4} + 6 A_{3}(\v_{1}, \v_{3}) + 3 A_{3} \v_{2}^{2} + 12 A_{4} (\v_{1}, \v_{1}, \v_{2})\nonumber\\& \quad\quad\quad\quad\quad+ 5 A_{5} \v_{1}^{4} - \eta^{(0)} \v_{4} - \eta^{(1)} \v_{3} - \eta^{(2)} \v_{2}- \eta^{(3)} \v_{1} =0 \nonumber 
\end{align}
By the first tangency equation, $\v_1$ is a unit eigenvector of $D^2 p(0)$ lying in the $\eta^{(0)}$-eigenspace. In many cases, but not always, tangency arcs are uniquely determined by $\v_1$.

\begin{prop}[Notation as above]\label{prop:poly_tan_nosym}
Suppose given $p\in P(\RR^d)_0$, $\deg p\ge3$. Let $\eta^{(0)}$ be an eigenvalue of $2A_2$, multiplicity $m$, let $Q\in M(d,m)$ be matrix whose columns constitute a basis for the associated eigenspace, and let $R$ complete the columns of $Q$ to a basis of eigenvectors for $\RR^d$ ($A_2$ symmetric). If $\balpha_0, \eta^{(1)}$ is such that  $3Q^\top A_3 (Q\balpha_0)^2 -  Q^\top Q\eta^{(1)}\balpha_0=0$ and $\x\mapsto (6Q^\top A_3 (Q\balpha_0, Q\x)-  \eta^{(1)}Q^\top Q\x)$ is invertible, then there exists a unique tangency arc such that
$\gamma(t) = tQ\balpha(t)+t^2 R\bbeta(t)$, $\balpha, \bbeta$ are real analytic in $t$, $\balpha(0) = \balpha_0$ and
\begin{align}
\bbeta(0) = -3(R^\top(2A_2 - \eta^{(0)}I_d)R)^{-1}R^\top A_3 (Q\balpha_0)^2.
\end{align}
If $m=1$, the condition simplifies to $\eta^{(1)} = A_3\v_1^3 \neq0$.
\end{prop}
\proof The proof proceeds by essentially following a Lyapunov-schmidt reduction and then invoking the real analytic implicit function theorem (e.g., \cite{Krantz1992}). We construct a real analytic $\gamma(t) = tQ\balpha(t)+t^2 R\bbeta(t)$ with $\balpha(0)=1$ (the parameterization forced by (\ref{tangency_eqns})). The equation defining $\balpha, \bbeta$ and $t$ is then
\begin{align*}
	\sum_{k=2}^{\deg(p)} kA_k(tQ\balpha+t^2R\bbeta)^{k-1}
	- (\eta^{(0)}+ t\eta^{(1)})(tQ\balpha+t^2R\bbeta) = 0.
\end{align*}
Since $Q\balpha$ is an $\eta^{(0)}$-eigenvalue, we may divide through by $t^2$ and so have
\begin{align}\label{prop:v one determines eqn2}
(2A_2 - \eta^{(0)}I_d)R\bbeta  + 3A_3 (Q\balpha)^2 -  \eta^{(1)}Q\balpha + O(t)  = 0.
\end{align}
We solve for $\bbeta$ at $\balpha = \balpha_0$ and $t=0$ by multiplying by $R^\top$ and observing that $R^\top(2A_2 - \eta^{(0)})R$ is invertible, and denote the solution by $\bbeta_0$. We may now regard the lhs of the above as a real analytic map $\Phi:\RR^m\times \RR^{d-m}\times\RR\to\RR^d$ of $(\balpha, \bbeta, t)$, and so find by our hypothesis on $\balpha_0$ and $\eta^{(0)}$ that $D_{(\balpha,\bbeta)}\Phi(\balpha_0,\bbeta_0,0)$  is non-singular, and so have $\balpha$ and $\bbeta$ as unique real analytic in $t$  by the real analytic implicit function theorem.
\qed\\

Besides their use in characterizing germs of tangency sets, tangency arcs also play an important role in relating the extremal characteristic of $f$ and $T^r_\c(f)$. For example, for a tangency arc $\gamma(t)$ constructed in the proof of \pref{prop:poly_tan_nosym},  $\gamma(t) = t\v + O(t^2)$, where $\v$ is an eigenvector of $\nabla ^2 f(0)$. Therefore,  if $\v$ is associated to a nonzero eigenvalue then, for sufficiently small $t>0$,  $f(\gamma(t))<0$ or $f(\gamma(t))<0$ iff $T_\c^2 f(\gamma(t))$ is. If the eigenvalue associated to $\v$ is zero, then higher-order derivatives may be required to draw a similar conclusion. Note, however, that analogously to finite determinacy of tangency sets and flat functions, a Taylor approximation determining the extremal characteristic may not always exist. We avoid the latter situation by placing generic conditions on jets, the degree of which depends on the application and the class of functions involved in the analysis, see \secsuf.

\subsection{Genericity, tensor eigenvalues and constructible sets} \label{sec: gen and const}
Referring to tangency equations, for every $k\ge2$, the $k$th tangency equation implies that $\v_1$ is an eigenvector of $kA_{k}$ modulo terms depending on lower-order tensors. The latter statement requires clarification as there is more than one way for  eigenvectors to generalize to tensors. Throughout, by eigenvectors we mean $E$-eigenvectors 
\cite{qi2005eigenvalues}. For  $\v_1$, being unit, this reads
\begin{equation}\label{tensor:eev}
	kA_{k} \v_1^{k-1} = \eta^{(k-2)}\v_1.
\end{equation}
In particular, $\eta^{(k-2)} = k A_{k} \v_1^{k}$. We say that $\eta^{(k-2)}$ is \emph{regular} if $0$ is a regular value of the map $\x\mapsto kA_{k} \x^{k-1} - \eta^{(k-2)}\x$ on the unit sphere. In particular, to any regular eigenvalue there correspond finitely many eigenvectors. 

We are interested in the situations where conditions equvalent to the hypotheses of \pref{prop:poly_tan_nosym} are generic.
\begin{definition}[\propC]
We say that $p\in P(\RR^d)_0$ satisfies \propC~if 
any eigenvalue of $D^3p(0)$ restricted to an eigenspace of $D^2p(0)$ is regular.
\end{definition}

\begin{rems}\label{rems: one dim v1}
1) A related approach, mentioned also in \pref{sec: ann}, is given by a powerful blowing-up construction used in \cite{buchner1983applications}. 
However, the critical assumption on the first nonvanishing homogeneous component being regular on its zero set does not hold in the intended applications. In addition, our goal is different in that we aim for isotopy equivalence.\\
2) Generalizations of \propC~to ones involving $\v_i,i>1$ and tangency equations to order $k$th, $k\ge3,$ are considered in \secsymres~and \cite{arjevani2025optsym}, and may also be formulated in terms of projective varieties intersecting transversely.
\end{rems}

Related to \propC, there is the question of estimating the number of tensor eigenvalues. The question has been addressed in \cite{cartwright2013number} by use of elementary properties of constructible sets and toric geometry. It is proved that if a tensor (not necessarily symmetric) has a finite number of eigenvalues, possibly complex, then their number counted with multiplicity is $((k-1)^d-1)/(k-2)$
(\emph{op. cit.}, Theorem 1.2). We repeat the part of the proof involving constructible sets as a way to introduce the topic (the preliminary section is quite long as it is). A~subset $\X$ of $\bC^n$ (or $\bP^n$) is constructible if it is a finite union of sets defined by conditions of the form $p=0$ and $q\neq0$, $p$ and $q$ polynomials. If $A$ is an order-$k$ tensor, not necessarily symmetric, the set of all eigenvalues can be given by the projection of the set comprising all eigenpairs $(\x,\eta)$, constructible by (\ref{tensor:eev}), on the $\eta$-coordinate and is therefore constructible by Chevalley's theorem. In particular, the set of tensor eigenvalues is either finite or cofinite. For symmetric tensors, the set of eigenvalues is exactly the set of critical values of $(\x\mapsto A\x^k)|_\sphere$, and is therefore finite being a set of measure zero by Morse-Sard's theorem. Consequently, the number of real $\eta^{(k-2)}$ satisfying (\pref{tensor:eev}) is \emph{always} bounded by $((k-1)^d-1)/(k-2)$ (\emph{op. cit.}, Theorem 5.5).
\begin{prop} \label{lem: E}
If $E$ is a subspace of $P^{k}(\RR^d)_0,k\ge3$, then the set $X$ of all $p\in E$ satisfying \propC~is open in $E$, and if $X$ is not empty then $X$ is dense in $E$.
\end{prop}
\proof 
Let $\E = E\otimes_\RR \CC\subseteq P^k(\CC^d)_0$ denote the complexification of $E$, and $\X$ the set of all $p\in \E$ satisfying \propC. We identify $E$ with the $\RR$-subspace $E\otimes_\RR 1\subseteq \E$. That $X$ and $\X$ are open follows for example by the number of symmetric tensor eigenvalues being finite, as well as standard results from (symmetric) matrix perturbation theory. 

To argue that $\X$ is constructible, we proceed similarly to the definable case expressing it in terms of first-order formulae ranging over constructible sets. We may restrict to orthogonal matrices without loss of generality. The set $\A_m$ of all pairs $(A,U) \in M_\CC(d,d)\times  M_\CC(d,m)$ such that $U^\top U = I_m$ (cf. Stiefel manifold) and the columns of $U$ form a basis of an $m$-dimensional eigenspace of $A$ is clearly constructible. Thus, $\X$ can be equivalently given as the set comprising all $p\in \E$ such that for any $m\in[d]$, if $(\nabla^2p(0),U)\in \A_m$ then for any $(\balpha, \eta)\in \CC^m\times \CC$: if $(\balpha, \eta)$ is an eigenpair of  $D^3 p(0)$ restricted by $U$, then $\eta$ is not a root of $\det [D(\x\mapsto D^3 p(0)\x^2)|_U(\balpha)-\lambda I_m]$. To prove density, observe that if $X$ is not empty, then $\X$ contains an interior point and so, being constructible, is dense in $\E$. Proceeding now as in \cite[Lemma A.6]{field1992symmetry}, since $\X$ is open, $\E\backslash \X$ is contained in an algebraic set of codimension 1. Therefore, by a well-known result (e.g., \cite{whitney1992elementary}) so is $E\backslash X$.
\qed\\

\begin{cor}\label{cor: e space}
The set of $p\in P^{3}(\RR^d)_0$ satisfying \propC~is dense and open.
\end{cor}
\proof 
By \pref{lem: E}, suffices it to show the existence of a single polynomial satisfying \propC. Take, for instance, $p\in P(\RR^d)_0$ defined by $$p(x_1,\dots,x_d) = \sum_{i=1}^d (ix_i^2/2 + x_i^3/3).$$ Then, the eigenspaces of $D^2p(0)$ are the coordinate axes $\RR\e_i$ and for every $i\in[d]$, $D^3p(0) \e_i^3 = 1\neq0$, concluding.
\qed\\

\begin{thm}[Classification of tangency set of $\gspace{e}$-germs]\label{thm: class_nosym}\phantom\newline
	
\noindent 
1) (Finite determinacy) $\tanset$-regular germs in $\gspace{e}$ are $3$-determined. \\
2) (Characterization) For a dense and open set of $\prescript{}{d}{\gspace{e}}$, the tangency set consists of $2d$ arcs approaching the origin along an eigenspace of~$D^2 f(0)$. \\
3) (\suf sufficient) The extremal type of $f$ at $0$ is determined by that of~$T^3_0 f$.
\end{thm}
\proof To prove the first claim, observe that whether $f = \sum_{i=1}^dh_i(\x)x_i^2 \in \fspace{}{U}$ is $\tanset$-regular depends on the values of $H(\x) = (h_1(\x),\dots,h_d(\x))$ and $DH(x)$ at the origin. The two quantities being determined by $T_0^3f$ implies that $f$ is $\tanset$-regular iff  $T_0^3f$ is, giving $3$-determinacy. For the second claim, we may assume
\propC, a generic condition by \pref{cor: e space}, and that all the $d$ Hessian eigenvalues are distinct, a generic condition as well (being defined by the nonvanishing of the discriminant of the characteristic polynomial of $D^2 p(0)$), and so use \pref{prop:poly_tan_nosym}. 
Lastly, if no eigenvalues vanishes, the third claim is a consequence of Morse lemma. Otherwise, use the condition on $D^3p(0)$ nonvanishing on the eigenspace associated to the zero eigenvalue.
\qed

% !TEX root = general_symmetry_and_critical_points.tex
\subsection{Tangency sets definable in o-minimal structures} \label{sec:omin_pre}
We cover generalities following directly from the theory of o-minimal structures, see \pref{sec: tan pre}, with emphasis on local results directed towards an o-minimal version of \cite{netto1984jet}. 

\begin{cor} \label{cor:finite_comp}
If $f$ is definable in some o-minimal structure $\cD$ and $\c$ a point, then $\tanset_{\c}(f)$ is definable. In particular, $\tanset_\c(f)$ has a finite number of definable $C^1$-piecewise path connected components.
\end{cor}

Sufficiently close to a critical point $\c$, the topological type of $\tanset_\c(f)\cap \sphere_\c(r)$  stabilizes and all critical points path connect to $\c$ through the tangency set, a consequence of the o-minimal theoretic local trivialization. Let $\ball_\c(r)$ denote the radius-$r$ ball centered at $\c$.

\begin{lemma}[Notation and assumptions as above]
	For $\epsilon>0$ sufficiently small,
	\begin{enumerate}[leftmargin=*]
		\item there exist definable connected manifolds $F_1,\dots, F_n\subseteq \RR^d$ such that $\tanset_\c\cap \mathring{\ball}_\c(\epsilon)$ is a finite union of mutually exclusive definable sets 
		%		$P_1,\dots,P_n\subseteq\RR^d$
		each homeomorphic to $(0,\epsilon)\times F_i$ for some~$i\in[n]$. 
		
		\item for each $i\in[n]$, $f$ is constant on $\sphere_\c(r)\cap F_i$ and is monotonic as a (now well-defined) function of the radius.
		
		\item if $\c'\in \mathring{\ball}_\c(\epsilon)$ is a critical point then $\c'$ and $\c$ lie in the same (path) connected component of the finitely many components of~$\tanset_\c(f)$.
	\end{enumerate}
\end{lemma}
\proof
Consider the Euclidean norm over the tangency set, $\|\cdot-\c\|:\tanset_\c\arr \RR$. By local triviality, there exists an interval $(0,\epsilon)$, a definable set $F$ and a definable homeomorphism $h:\tanset_\c\cap \mathring{\ball_\c}(\epsilon)\to (0,\epsilon)\times F$. Writing $F$ as a union of definable connected sets gives (1) (more structure may be deduced by using a local trivialization \emph{respecting} $\tanset_\c(f)$, implying that $\tanset_\c(f)$ is in fact locally homeomorphic to a cone). To show (2), assume first that two points $\x,\y$ on $\sphere_\c(r)\cap F_i$ are connected by a $C^1$-path $\gamma$, having $\gamma(0) = \x$ and $\gamma(1)=\y$. Then,
\begin{align*}
f(\y) &=  \int_{0}^{1} \inner{\nabla f(\gamma(t)), \dot{\gamma}(t)} dt + f(\x)\\
&= \int_{0}^{1} \inner{\eta(t)(\gamma(t)-\c), \dot{\gamma}(t)} dt + f(\y)
=  f(\x),
\end{align*}
where $\eta(t)$ is defined by $\grad f(\gamma(t))=\eta(t)(\gamma(t)-\c)$ and the second equality follows by $\|\gamma(t)-\c\|=r$. Now use \pref{cor:finite_comp}. The monotonicity of $f$ regarded as a function of the radius for $\epsilon>0$ sufficiently small is a consequence of the monotonicity theorem, concluding~(2). Finally, since $\critset\subseteq\tanset_\c(f)$, if $\c'$ is a critical point at distance at most $\epsilon$ from $\c$ then $\c' = h^{-1}(\epsilon',\x')$ for some $\epsilon'\in(0,\epsilon)$ and $\x'\in F_i$, and we may construct a tangency arc connecting $\c$ to $\c'$ by defining $\gamma(t) \defoo h^{-1}(t\epsilon', \x)$ $(\gamma(0) \defoo \c)$. \qed\\

Recall that if $f$ is definable then $\minimizers$, the set of minimizers of $f$ restricted to $\sphere_{\c}(r)$ (see (\ref{minimizers})), is also definable. Thus, by the CSL, we may construct a continuous definable arc $\gamma$  in $\minimizers$ parameterized by arc length. The following simple corollary is key to our study of \suf-sufficiency.
\begin{cor} \label{cor:mr_exists}
If $f$ is definable, then there exists a tangency arc $\gamma$ parameterized by arc length satisfying $\ploss(\gamma(r)) = m_\c(r)$, and similarly for~$M_\c(r)$.
\end{cor}
Despite the simplicity of the proof, the existence of an arc giving $m_\c(r)$ is not obvious. Counter-examples exist already for (necessarily non-definable) functions on $\RR$, e.g., $x\mapsto \mathrm{ReLU}^7(x)\sin(1/x)$.  

If $f$ is definable in an o-minimal structure $\cD$, the existence of tangency arcs in \pref{cor:mr_exists} describing the minimum and the maximum of $f$ over $\sphere_\c(r), r>0$, implies that the extremal type of $\c$ can be completely determined by tangency arcs. If $\cD$ is further polynomially bounded then by growth dichotomy $f(\gamma(t)) = a t^\alpha + o(t^\alpha)$ for some $\alpha\in\RR, a\in\RR\backslash\{0\}$.

% !TEX root = general_symmetry_and_critical_points.tex

\section{The symmetric tangency set}
\label{sec:sym_tan}

\subsection{Preliminaries} \label{sec:pre_sym}
In the sequel, elementary properties of group actions are assumed known.

\subsubsection{Representations} \label{sec: reps}
Let $V$ be a finite dimensional real vector space with inner product $\tri{\,,\,}$ and norm $\|\;\|$. An $\RR$-representation of a finite group $G$ is a group homomorphism of $G$ to $\GL{V}$, the general linear group on $V$. 
The \emph{degree} of the representation is the dimension of $V$. If $G$ acts by permuting the elements of some basis elements of $V$, we say that $(V, G)$ is a (linear) \emph{permutation representation}.
If $G$ acts orthogonally on $V$, we say that $(V,G)$ is an  \emph{orthogonal representation}. Usually, we just say $(V,G)$ is a \emph{representation} of $G$, omitting the indication of the underlying field being $\RR$ and assume orthogonality.  

A representation $(V, G)$ is \emph{trivial} if each element of $G$ acts as the identity on $V$. If $(V_1,G)$ and $(V_2,G)$ are two representations, the direct sum $V_1\oplus V_2$ and the tensor product $V_1 \otimes V_2$ are also representations, the latter via $g(\v_1\otimes \v_2) = g\v_1\otimes g\v_2$. The $m$th tensor power $V^{\otimes m}$, the exterior powers $\wedge^m V$ and the symmetric powers $\symp^m(V)$ are representations by the same rule. The dual representation $V^* = \hom(V, \bR)$ is defined by $g\v^* = \v^*\circ g^{-1}, \v^* \in V^*$, making $\hom(V_1, V_2)$ a representation via the identification $\hom(V_1, V_2) = V_1^*\otimes V_2$. If $H\subseteq G$ is a subgroup, any representation $(V, G)$ restricts to a representation of $H$, denoted by $\mathrm{Res}_H^G V$. If 
$(W,H)$ is a representation of $H$, the induced representation is $\mathrm{Ind}_H^G W = \RR[G]\otimes_{\RR[H]}W$.

A map $f: V_1 \arr V_2$ is called $G$-equivariant if $f(gv) = gf(v)$, for all $g \in G, v \in V_1$. If $G$ acts \emph{trivially} on $V_2$, then $f$ is said to be $G$-invariant. An equivariant linear map $A$ is a \emph{$G$-map}.  The gradient field of a $G$-invariant function is a $G$-equivariant self map of $V$. Corresponding to each subgroup $H\subseteq G$ is the fixed point linear subspace defined by $V^H \defoo \{\x \in V \dd g\x = \x, \forall h\in H\}$. By equivariance, $\nabla f$ restricts to any fixed point space. Thus, if $\c$ is a critical point of a $G$-invariant function $f$ then $\nabla f(\c) \in V^{G_\c}$ and  $\v\mapsto \nabla^2f(\c)\v$ is a $G_\c$-self map of $V$.

A representation is \emph{irreducible} if there are no proper $G$-invariant subspaces of $V$. If $A$ is a $G$-map then $\lker(A)$ and $\im(A)$ are $G$-invariant linear subspaces of $V$. Consequently, if $A:V \arr V$ is a $G$-map and $(V,G)$ is irreducible, then $A$ is either the zero map or a linear isomorphism. The orthogonal complement of $\im(A)$ is $G$-invariant and so if $A \ne 0$, then $A$ must be onto; a similar argument using $\lker(A)$ shows $A$ is 1:1. If $(V,G)$ is irreducible, then the space $\text{Hom}_G(V,V)$ of $G$-maps (endomorphisms) of $V$ is a real associative division algebra and is isomorphic by a theorem of Frobenius to either $\real, \mathbb{C}$ or $\mathbb{H}$ (the quaternions). The \emph{only} case that will concern us here is when $\text{Hom}_G(V,V) \approx \real$, where we say that the representation is \emph{absolutely irreducible}. 

The character of a representation $(V, G)$ is the complex-valued function $\chi_{V}: V\to\bC$ defined by $\chi_{V}(g) = \tr(g:V\mapsto V)$. Thus, for example, if $(V, G)$ is a permutation representation then $\chi_V(g)$ is the number of basis elements fixed by $g\in G$. 
The inner product of two characters $\chi_{V_1}$ and $\chi_{V_1}$, being instances of class functions over $G$, is given by $\inner{\chi_{V_1}, \chi_{V_2}} \defeq \frac{1}{|G|}\sum_{g\in G} \overline{\chi_{V_1}(g)}\chi_{V_2}(g)$. By a well-known result, if $V_1$ is irreducible then its multiplicity  in a representation $V$ is $\inner{\chi_V, \chi_{V_1}}$. In particular, the isomorphism class of a representation is determined by its character. The degree of $V$ may be computed by evaluating $\chi_V$ on the  identity.

\subsubsection{Irreducible representations of the symmetric group}
Each irreducible representation of $S_d$ is associated to a partition $\lambda = (\lambda_1,\dots,\lambda_m)$, $\lambda_1 \ge \dots \ge \lambda_m \ge0$ of~$d$. We write $\lambda \vdash d$ for short and denote the associated representation by $V_\lambda$. When no confusion results, we often let $V_\lambda^{\oplus m}$ denote a concrete realization of the isotypic component in a given representation, rather than just the isomorphism class.

\begin{exam}
Referring to (\ref{iso_decomp_sym1}), the representation $(\RR^d,S_d)$ decomposes into two isotypic components given by: the trivial representation $V_{(d)}$ and the \emph{standard} representation $V_{(d-1, 1)}$, $d \ge 2$, on $\{\x \in \real^d~|~\sum_{i\in\is{d}}x_i = 0\}\subset \real^{d}$. 
\end{exam}

Irreducible characters $\chi_\lambda$ may be evaluated using Frobenius' character formula. If $C_{\i}$ is a conjugacy class of $S_d$ corresponding to a \emph{cycle type} $\i = (i_1,i_2\dots,i_d)$,
$i_j$ giving the multiplicity of cycles of length $j$ and 
$i_1 + 2i_2 +\dots + di_d = d$, then
\begin{align}\label{formula:fro}
	\chi_\lambda(C_\i) = \brk*{\Delta(x)\prod_{j} p_j(x)^{i_j}}_{\lambda_1 + d - 1, \lambda_2 + d - 2, \dots, \lambda_m}
\end{align}
where $\Delta(x)=\prod_{i<j}(x_i-x_j)$ is the discriminant and $[r(x)]_{l_1\dots, l_k}$ generally denotes the coefficient of $x_1^{l_1}\cdots x_k^{l_k}$ in a formal power series~$r$.

\subsubsection{Invariant theory}
Let $(V, G)$ be a representation. By the preceding section, the (linear) action of $G$ extends to the graded algebra of polynomial functions $P(V)$ by the identification with $\symp(V^*)$, the symmetric algebra of $V^*$. The {homogeneous components} are given by $P^k(V)\defeq \symp^k(V^*)$. If $\v^*_1,\dots, \v^*_m\in V^*$ is a basis, $P(V)$ may be equivalently given by $\bR[\v^*_1, \dots, \v^*_m]$. The \emph{ring} or \emph{algebra of invariants}, denoted by $P(V)^G$, is the fixed subalgebra,
\begin{align}
	P(V)^G = \{ f\in P(V)~|~ gf = f, \text{for all } g\in G\}.
\end{align}
If $(V_1, G)$ and $(V_2, G)$ are representations, the space of $G$-equivariants polynomial maps $P(V_1,V_2)^G \defeq (\symp(V_1^*) \otimes V_2)^G$ is a graded module over $P(V_1)^G$. The homogeneous components are given by $P^k(V_1, V_2)^G \defeq (\symp^k(V_1^*) \otimes V_2)^G$. If $U\subseteq V_1$ is an open set, the ring of invariant smooth functions $C^\infty(V_1)^G$ and the $C^\infty(V_1)^G$-module of equivariant smooth maps $C^\infty(V_1,V_2)^G$, as well as maps with finite differentiability order, are likewise~defined. As earlier, adding the subscript $\c$ in the latter indicates that at $\c$, the value and the first-order derivatives vanish. Finally, we let $\gspace{G}$ denote the ring of invariant smooth germs of $f\in C^\infty(V_1)^G_0$ at $\c=0$. 

 Methods for computing homogeneous invariants vary by the nature of the representation considered, e.g., \cite{derksen2015computational}. For permutation representation, a natural basis of invariants for $P_G^k(V)$ is given by \emph{orbits sums}, i.e., sums of orbit elements, where an orbit is defined by 
\begin{align}\label{orbit}
Gx_1^{i_1}\cdots x_k^{i_k} \defoo \{ gx_1^{i_1}\cdots x_k^{i_k}~|~g\in G\}.
\end{align}
with the induced action  on $P^k(V)$.

By a theorem of Hilbert, $P(V)^G$ is a Noetherian ring (polynomial exactly when $G$ is generated by pseudoreflections). By a result of Noether, there exists a generating set consisting of homogeneous invariants whose degree is less than or equal to $|G|$. For permutation representations, a generally tighter bound exists.
\begin{thm}[\cite{garsia1984group}]\label{thm: bound gens sd}
If $(V,G)$ is a permutation representation, then there exists a generating set for $P(V)^G$ consisting of homogeneous invariants whose degree is bounded by $\max\{\dim V, \binom{\dim V}{2}\}$. 
\end{thm}
The bound was proved in \cite{garsia1984group} for the nonmodular case and is sharp as seen by $\QQ[x_1,\dots,x_d]^{A_d}$, $A_d$ being the alternating group, tautological representation. An explicit construction holding in arbitrary characteristic was given later by \cite{gobel1995computing}.

\subsection{Regularity and equivariant jet transversality}
Just as our definition of regular tangency sets is formulated in terms of jet transversality, regularity in the symmetric case is formulated in terms of \emph{equivariant} jet transversality. The extension of Thom's jet transversality to the category of equivariant maps is due to Bierstone \cite{bierstone1976generic} (equivariant transversality of maps was introduced earlier and is due to Bierstone \cite{bierstone1977general} and Field \cite{field1977transversality}). 

The definition of $\tanset$-regularity (\pref{def: reg}) follows by first expressing $f\in \gspace{e}$ in terms of the set of  $P(\RR^d)_0$-generators $\{x_i^2\}$, i.e., $f(\x) = \sum_{i=1}^d h_i(\x)x_i^2$, $H=(h_1,\dots,h_d)$ smooth, and then imposing a transversality condition on $\{h_i\}$. The definition carries over to the symmetric case by simply taking generators of $P(V)_0^G$ instead. 

In the general theory, one considers two representations
$(V,G)$ and $(W,G)$, and minimal sets of generators:
$\{p_1,\dots, p_l\}$ for $P(V)^G$ and $\{F_1,\dots,F_k\}$ for the $P(V)^G$-module $P(V,W)^G$. By the work of Schwarz and Malgrange, $f\in C^\infty(V,W)^G$ may be written as 
\begin{align} \label{rep f schwarz}
f = \sum_{i=1}^k (h_i\circ P)F_i,
\end{align}
where $P = (p_1,\dots,p_l)$ is the \emph{orbit map} and $H=(h_1,\dots,h_k)$  is a smooth map. Thus, $j^kf$ factors as 
\begin{align} \label{rep jf schwarz}
j^q f = u_{P}^q\circ (I, H^q\circ P),
\end{align}
where $H^q(\x) = (H(\x), DH(\x), \cdots, D^qH(\x))$ and $u_P^q:V\times P^q(\RR^l,\RR^k)\to J^q(V,\RR)$ is an equivariant polynomial map depending on the $q$-jets of~$P$ and $F_i$. Equivariant jet transversality to a $G$-invariant closed semialgebraic subset of $J^q(V,\RR)$, written $j^qf \trans_G Q$, is then defined 
by requiring $(I, H \circ P)\trans_{G} (u_P^q)^{-1}(Q)$, with respect to the canonical Whitney stratification. Regularity of tangency sets of invariant functions is now defined by replacing $e$ with $G$ in \pref{def: reg} with the additional modifications necessary  to account for invariant functions where both their value and first-order derivatives vanish at a fixed point. We write $(V,G)$, rather than $G$, when we want to ensure that no ambiguity results.

\begin{definition} \label{def: greg}
Let $(V,G)$ be a representation and let $\uniset_{(V,G)}$ denote the universal stratification. Given $f\in \fspacec{(V,G)}{U}{\c}$, if $j^1 (I, H \circ P)\trans_{(V,G)} \uniset_{(V,G)}$, we say that $\tanset_{\c}(f)$ is regular and that $f$ is $\tanset_{\c}$-regular. Regularity of the germ of $\tanset_{\c}(f)$ at $\c$ is likewise defined using the function germ. In both cases, if $\c=0$, we usually simply write $\tanset$- rather than $\tanset_\c$-regular. Finite determinacy extends to the equivariant category mutatis mutandis.
\end{definition}
By Theorems 7.6, 7.7, and 7.8 in \cite{bierstone1976generic}, we find that,
\begin{cor}
openness, density, and  isotopy theorems hold for $\tanset$-regularity in the equivariant category.
\end{cor}

\begin{prop}\label{prop: fd of permreps}
Let $(V,G)$ be a permutation representation. If $f\in \gspace{G}$ is $\tanset_{\c}$-regular, then $\tanset_{\c} (f)$ is finitely determined.
\end{prop}
\proof
By \pref{thm: bound gens sd}, there exists a minimal generating set  $\{p_1,\dots, p_l\}$ for $P(V)^G$ with $l\le \max\{\dim V, \binom{\dim V}{2}\}$. Proceed now along the lines of the proof of \pref{thm: class_nosym}.
\qed\\

\begin{rem}
A significant part of the development in \cite{bierstone1976generic} is dedicated to demonstrating that equivariant jet transversality is well-defined, i.e., independent of all choices. That $p_i$ and $F_i$ appearing in the representation (\ref{rep f schwarz}) can be used to deduce $(\max \deg p_i + \max \deg F_i)$-determinacy is due \cite{field1989equivariant}.
\end{rem}
\subsection{Stabilizing characters of tensor powers of $(\RR^d, S_d)$} \label{sec: stab chars}
Our study of functions invariant under permutation representation requires the knowledge of the isotypic decompositions of certain tensor and symmetric powers of $V_\lambda$. Frobenius' formula (\ref{formula:fro}) gives a systematic way to compute such decompositions for any arbitrarily large dimensionality. For instance, to establish (\ref{decomp_mdd}) for $d\ge 4$, observe that by Frobenius' formula (\ref{formula:fro}), 
\begin{align}\label{irr_char}
	\chi_{(d-1,1)}(C_\i) &= i_1 - 1,\\
	\chi_{(d-2,1, 1)}(C_\i) &= \frac{1}{2}(i_1 - 1)(i_1 - 2) - i_2,\nonumber\\ 
	\chi_{(d-2, 2)}(C_\i) &= \frac{1}{2}(i_1 - 1)(i_1 - 2) + i_2 - 1,\nonumber
\end{align}
where $\i = (i_1,i_2\dots,i_d)$ denotes a cycle type. Computing, we find that
\begin{align*}
\chi_{(\RR^d)^{\otimes 2}}(C_\i) &= \chi_{\RR^d}(C_\i)^2 = i_1^2 \\&= 2\chi_{(d)}(C_\i) + 3\chi_{(d-1, 1)}(C_\i) + \chi_{(d-2, 1, 1)}(C_\i) + \chi_{(d-2, 2)}(C_\i).
\end{align*}
Given a \emph{fixed} partition $\alpha = (\alpha_1,\cdots,\alpha_m), \alpha_1\ge\cdots \ge\alpha_m\ge0,$ independent of $d$, we define $$\alpha[d] \defeq (d- |\alpha|,\alpha_1, \alpha_2, \dots, \alpha_m),$$
where $|\alpha| \defeq  \alpha_1+\cdots+\alpha_m$. As suggested by the previous example, for any such $\alpha$, $\chi_{\alpha[d]}$ is  polynomial in $i_1,\dots, i_d$. Characters of multilinear algebraic constructions involving representations of the form $V_{\alpha[d]}$ are also polynomials in $i_1,\dots,i_d$ and can therefore be given as linear combinations of $\chi_{\alpha[d]}$ from which the desired isotypic decomposition may  be deduced. The approach, presented in \cite{murnaghan1938analysis}, proceeds by establishing bounds on $w(p)$, the \emph{weight} of a polynomial $p$ in $i_1,\dots,i_d$, giving the maximal weighted degree $j_1+\cdots+d j_d$ on its terms $c_{j_1,\dots, j_d}i_1^{j_1}\cdots i_d^{j_d}$. 

\begin{prop}[\cite{murnaghan1938analysis}. Notation as above]
Any polynomial $p$ in $i_1,\dots,i_d$ with $w(p)\le d$ is a linear combination of irreducible characters of the form $\chi_{\alpha[d]}$ with $|\alpha| \le w(p)$. Moreover, for any fixed partition $\alpha$, $w(\chi_{\alpha[d]})\le |\alpha|$.
\end{prop}
For example, the irreducible characters involved in describing $\chi_{(\RR^d)^{\otimes 2}}$ correspond exactly to partitions $\alpha$ with $|\alpha|\le w(\chi_{(\RR^d)^{\otimes 2}}) = 2$, namely, $(), (1), (2), (1,1)$. The coefficients in the decomposition stabilize when all $\alpha[d]$ become valid partitions of $d$, namely, when $d - |\alpha|\ge  \alpha_1$ (hence, for $(\RR^d)^{\otimes 2}$, $d\ge4$). 
For symmetric powers of $\RR^d$, we obtain (\ref{iso_decomp_sym2}) and 
(\ref{iso_decomp_sym3}) using (\ref{irr_char}) and the identity
\begin{align*}
\chi_{(d-3, 3)}(C_\i) &= \frac{i_1(i_1-1)(i_1-5)}{6} +i_2(i_1-1) + i_3.
\end{align*}

\subsubsection{Homogeneous invariants of $(\RR^d, S_d)$} The orbit sum basis  of $P^k(\RR^d)$ is given by the monomial symmetric polynomials $m_\lambda(x) = \sum_{\mu} x^\mu$, the sum ranging over all distinct permutations $\mu$ of a partition $\lambda \vdash k$. For $d\ge 3$, we have 

\begin{align}\label{sd_hom_invs}
	P^2(\RR^d) &= \mathrm{Span}\{ m_{(2)}, m_{(1, 1)}\},\nonumber\\
	P^3(\RR^d) &= \mathrm{Span}\{ m_{(3)}, m_{(2, 1)}, m_{(1,1,1)}\}.
\end{align}
That $\dim P^2(\RR^d) = 2$ and $\dim P^3(\RR^d) = 3$ for any sufficiently large $d$ can also be deduced from (\ref{iso_decomp_sym2}) and (\ref{iso_decomp_sym3}) by reading off the multiplicity of the trivial representation $V_{(d)}$. The method generally enables the computation of $\dim P^k(V)^G$ for an infinite sequences of representations of the form $\alpha[d]$, $\alpha$ a fixed partition. For example,
\begin{align}\label{msdsd_qinvs}
\chi_{\symp^2((\RR^{d})^{\otimes 2})}&= 
11 \chi_{(d)}
+ 21 \chi_{(d - 1 ,1 )}
+ 19 \chi_{(d - 2 ,2 )}
+ 13 \chi_{(d - 2 ,1 ,1 )}
+ 6 \chi_{(d - 3 ,3 )}\nonumber\\
&+ 10 \chi_{(d - 3 ,2 ,1 )}
+ 4 \chi_{(d - 3 ,1 ,1 ,1 )}
+ \chi_{(d - 4 ,4 )}
+ \chi_{(d - 4 ,3 ,1 )}
+ 2 \chi_{(d - 4 ,2 ,2 )}\nonumber\\
&+ \chi_{(d - 4 ,2 ,1 ,1 )}
+ \chi_{(d - 4 ,1 ,1 ,1 ,1 )},
\end{align}
\begin{align}\label{msdsd_cinvs}
\chi_{\symp^3((\RR^{d})^{\otimes 2})}&=
52 \chi_{(d)}
+ 147 \chi_{(d - 1 ,1 )}
+ 175 \chi_{(d - 2 ,2 )}
+ 157 \chi_{(d - 2 ,1 ,1 )}\nonumber\\
&+ 111 \chi_{(d - 3 ,3 )}
+ 183 \chi_{(d - 3 ,2 ,1 )}
+ 79 \chi_{(d - 3 ,1 ,1 ,1 )}
+ 36 \chi_{(d - 4 ,4 )}\nonumber\\
&+ 86 \chi_{(d - 4 ,3 ,1 )}
+ 54 \chi_{(d - 4 ,2 ,2 )}
+ 70 \chi_{(d - 4 ,2 ,1 ,1 )}
+ 20 \chi_{(d - 4 ,1 ,1 ,1 ,1 )}\nonumber\\
&+ 6 \chi_{(d - 5 ,5 )}
+ 16 \chi_{(d - 5 ,4 ,1 )}
+ 20 \chi_{(d - 5 ,3 ,2 )}
+ 18 \chi_{(d - 5 ,3 ,1 ,1 )}\nonumber\\
&+ 18 \chi_{(d - 5 ,2 ,2 ,1 )}
+ 12 \chi_{(d - 5 ,2 ,1 ,1 ,1 )}
+ 4 \chi_{(d - 5 ,1 ,1 ,1 ,1 ,1 )}
+ \chi_{(d - 6 ,6 )}\nonumber\\
&+ \chi_{(d - 6 ,5 ,1 )}
+ 2 \chi_{(d - 6 ,4 ,2 )}
+ \chi_{(d - 6 ,4 ,1 ,1 )}
+ 2 \chi_{(d - 6 ,3 ,3 )}\nonumber\\	
&+ 2 \chi_{(d - 6 ,3 ,2 ,1 )}
+ \chi_{(d - 6 ,3 ,1 ,1 ,1 )}
+ 2 \chi_{(d - 6 ,2 ,2 ,2 )}
+ 2 \chi_{(d - 6 ,2 ,2 ,1 ,1 )}\nonumber\\
&+ \chi_{(d - 6 ,2 ,1 ,1 ,1 ,1 )}
+ \chi_{(d - 6 ,1 ,1 ,1 ,1 ,1 ,1 )}.
\end{align}
Therefore, for sufficiently large $d$, $\dim P^2((\RR^d)^{\otimes 2}) = 11$ and $\dim P^3((\RR^d)^{\otimes 2}) = 52$. Some of the character formulae that we use for computing the isotypic decomposition of $\symp^2((\RR^{d})^{\otimes 2})$ and $\symp^2((\RR^{d})^{\otimes 3})$ are due to Frobenius \cite{frobenius1900charaktere} and Murnaghan~\cite{murnaghan1937characters}, for example, 
\begin{align*}
\chi_{(d - 3, 2, 1)}(C_\i) &= \frac{i_{1} \left(i_{1} - 4\right) \left(i_{1} - 2\right)}{3} - i_{3},\\
\chi_{(d - 3, 1, 1, 1)}(C_\i) &= - i_{2} \left(i_{1} - 1\right) + i_{3} + \frac{(i_{1}-1)\left(i_{1} - 3\right) \left(i_{1} - 2\right)}{6},\\
\chi_{(d - 4, 4)}(C_\i) &= \frac{i_{1} i_{2} \left(i_{1} - 3\right)}{2} + \frac{i_{1} \left(i_{1} - 7\right) \left(i_{1} - 2\right) \left(i_{1} - 1\right)}{24} + \frac{i_{2} \left(i_{2} - 1\right)}{2} \\&+ i_{3} \left(i_{1} - 1\right) + i_{4},\\
\chi_{(d - 4, 3, 1)}(C_\i) &= \frac{i_{1} i_{2} \left(i_{1} - 3\right)}{2} + \frac{i_{1} \left(i_{1} - 6\right) \left(i_{1} - 3\right) \left(i_{1} - 1\right)}{8} - \frac{i_{2} \left(i_{2} - 3\right)}{2} - i_{4},\\
\chi_{(d - 4, 2, 2)}(C_\i) &= \frac{i_{1} \left(i_{1} - 5\right) \left(i_{1} - 4\right) \left(i_{1} - 1\right)}{12} + i_{2} \left(i_{2} - 2\right) - i_{3} \left(i_{1} - 1\right),\\
\chi_{(d - 4, 2, 1, 1)}(C_\i) &= - \frac{i_{1} i_{2} \left(i_{1} - 3\right)}{2} + \frac{i_{1} \left(i_{1} - 5\right) \left(i_{1} - 3\right) \left(i_{1} - 2\right)}{8} - \frac{i_{2} \left(i_{2} - 1\right)}{2} + i_{4},\\
\chi_{(d - 4, 1, 1, 1, 1)}(C_\i) &= - \frac{i_{2} (i_{1}-1)\left(i_{1} - 2\right) }{2}  + \frac{i_{2} \left(i_{2} - 1\right)}{2} + i_{3} \left(i_{1} - 1\right) - i_{4} \\&+ \frac{(i_{1}-1)\left(i_{1} - 4\right) \left(i_{1} - 3\right) \left(i_{1} - 2\right)}{24}.
\end{align*}

% !TEX root = general_symmetry_and_critical_points.tex

\subsection{Tangency sets of $(\RR^d, S_d)$-invariant functions} \label{sec: tan sd}
We begin by proving the genericity of \propC. When $d$ is even, establishing uniqueness for arcs having isotropy $S_{d/2}\times S_{d/2}$ requires higher-order versions of \propC~which we provide in \secsymres.
\begin{lemma}\label{lem: propc in sd}
If $d$ is odd then the set of $p\in P^m(\RR^d)_0^{(\RR^d, S_d)},m\ge3$  satisfying \propC~is open and dense.
\end{lemma}
\proof 
By \pref{lem: E}, it suffices to demonstrate the existence of a 
$p\in P^m(\RR^d)_0^{(\RR^d, S_d)}$ satisfying \propC. Define
\begin{align}\label{hd propc}
h_d(\x) = \sum_{i<j} x_ix_j + \frac{1}{3}\sum_{i=1}^d x_i^3.
\end{align}
The polar forms of $D^2h_d(0)$ and $D^3h_d(0)$ are $A_2(\x,\y) = \frac{1}{2}\sum_{i,j=1}^d  x_iy_j$ and $A_3(\x,\y,\z) = \frac{1}{3}\sum_{i=1}^d x_iy_iz_i$, respectively. It can now be verified that \propC~holds for $h_d$ by straightforward computation. \qed\\
\begin{thm}[Classification of $(\RR^d, S_d)$-tangency sets]\label{thm: class_sd}
Generically, the number of tangency arcs is at most exponential in $d$, and the only admissible isotropy group are ones conjugated to $S_{d-q}\times S_{q}, 0\le q\le \floor{d/2}$.
\end{thm}
\proof
If $f$ is a smooth $S_d$-invariant function, the second- and third-order forms are given by,
\begin{align*}
A_2\x^2 &= a_1 M_{(2)}\x^2 + a_2 M_{(1, 1)}\x^2,\\
A_3\x^3 &= a_3 M_{(3)}\x^3 + a_4 M_{(2, 1)}\x^3 + a_5 M_{(1, 1, 1)}\x^3, 
\end{align*}
where $M_{\lambda}$ are the polar forms of $m_\lambda$
given in (\ref{sd_hom_invs}). The two eigenvalues 
of $2A_2$ corresponding to $V_{(d)}$ and $V_{(d-1,1)}$ are $2a_1 + (d-1)a_2$ and $2a_1-a_2$, respectively, and may be assumed distinct by the generic condition $a_2\neq 0$.  Since $\x\mapsto 3A_3\x^2-\eta\x$ is $G$-equivariant for any $G = S_{d-q}\times S_{q}, q\in \{0,\dots,\floor{d/2}\}$ (and conjugated subgroups), the equations defining the eigenvectors of $3A_3$ restrict to the corresponding $G$-fixed point space. Moreover, the restrictions to these fixed point spaces preserve the degree of the equations generically over the choice of $a_3,a_4$ and $a_5$. We may now assume by \pref{lem: propc in sd} that $f$ satisfies \propC~(see the discussion preceding \pref{lem: propc in sd}), and so have a finite number of solutions on $V_{(d)}$ and $V_{(d-1,1)}$. Consequently, counting, we find by Bezout's theorem that the solutions on the  fixed point spaces intersected with $V_{(d)}$ and $V_{(d-1,1)}$ exhaust all the solutions in $V_{(d)}$ and $V_{(d-1,1)}$ (note that all the arguments involved in the development of \propC~hold over~$\CC$). We may now use \pref{prop:poly_tan_nosym} and show that these account for all tangency arcs, concluding.
\qed\\

\begin{rems}
1) Generally speaking, a tangency set $\tanset_{\c}(f)$ may be equivalently formulated as the set of all pairs  $\{(\x,\lambda)~|~\nabla f(\x) = \eta (\x-\c)\}$ projected onto the $\x$-axis. Accordingly, we may regard  tangency arcs as the projection of solution curves bifurcating from the trivial solution $t \mapsto (\c,t)$ (compare to \cite{arjevanifield2022equivariant}), $\eta$ being the bifurcation parameter and the Hessian eigenvalues being the  bifurcation points. This allows for  the use of techniques from bifurcation theory, such as  the equivariant version of the center manifold theorem \cite{ruelle1973bifurcations} which may be used to locally reduce the analysis to irreducible representations.
\\
2) In the context of the irreducible standard representation $(V_{(d-1, 1)}, S_d)$, a study of the steady state bifurcation of $(V_{(d-1, 1)}, S_d)$-equivariant vector fields was carried out by~\cite{field1992symmetry2}. Our results differ in that they concern genericity of \emph{gradient} vector fields relative to the defining representation $(\RR^d, S_d)$, reducible, and emphasize connectedness to adjacent critical points. 
\end{rems}

\subsection{Tangency sets of $(M(d,d), S_d)$-invariant functions} \label{sec: tan dsd}
We describe several admissible isotropy groups occurring for $(M(d,d), S_d)$-tangency sets. More detailed statements and the complete classification theorem are deferred to \secsymres~after the relevant group-theoretic notions and general results concerning dimensionality of fixed point spaces have been introduced.

By (\ref{msdsd_qinvs}), the space of quadratic $(M(d,d), S_d)$-invariants is 11-dimensional and is spanned by the orbit sums of:
\begin{align*}
&\{ S_d w_{11}w_{23},
 S_d w_{12}w_{21},
 S_d w_{12}^2,
 S_d w_{11}w_{21},
 S_d w_{11}^2,
 S_d w_{12}w_{23},\\&
 S_d w_{11}w_{12},
 S_d w_{12}w_{32},
 S_d w_{11}w_{22},
 S_d w_{12}w_{34},
 S_d w_{12}w_{13}\}.
\end{align*}
By (\ref{msdsd_cinvs}), the space of cubic $(M(d,d), S_d)$-invariants is 52-dimensional and is spanned by the orbits sums of:
\begin{align*}
&\{S_d w_{11}w_{23}w_{45},
S_d w_{11}w_{23}w_{24},
S_d w_{11}w_{12}w_{33},
S_d w_{12}w_{21}w_{31},
S_d w_{11}w_{12}w_{22},\\&
S_d w_{12}w_{13}w_{42},
S_d w_{11}w_{21}w_{31},
S_d w_{11}w_{12}w_{13},
S_d w_{12}w_{23}w_{45},
S_d w_{12}^2w_{32},\\&
S_d w_{12}w_{13}w_{41},
S_d w_{11}w_{12}w_{31},
S_d w_{12}^2w_{34},
S_d w_{12}^2w_{31},
S_d w_{11}w_{21}^2,
S_d w_{12}w_{32}w_{45},\\&
S_d w_{12}w_{21}w_{34},
S_d w_{12}w_{23}w_{42},
S_d w_{11}w_{23}w_{43},
S_d w_{11}w_{21}w_{34},
S_d w_{11}w_{21}w_{32},\\&
S_d w_{11}w_{12}w_{34},
S_d w_{11}w_{12}w_{32},
S_d w_{11}w_{12}^2,
S_d w_{12}w_{23}w_{43},
S_d w_{11}^2w_{23},
S_d w_{12}w_{34}w_{56},\\&
S_d w_{12}^2w_{21},
S_d w_{11}^2w_{22},
S_d w_{11}w_{21}w_{23},
S_d w_{12}w_{32}w_{42},
S_d w_{11}w_{12}w_{23},
S_d w_{11}w_{12}w_{21},\\&
S_d w_{11}w_{22}w_{33},
S_d w_{12}w_{23}w_{31},
S_d w_{11}w_{23}w_{34},
S_d w_{11}w_{23}w_{32},
S_d w_{11}^3,
S_d w_{12}w_{13}w_{21},\\
&S_d w_{12}^2w_{23},
S_d w_{11}w_{23}^2,
S_d w_{12}w_{13}w_{23},
S_d w_{12}^2w_{13},
S_d w_{12}w_{23}w_{34},
S_d w_{12}^3,
S_d w_{11}^2w_{21},\\&
S_d w_{12}w_{13}w_{45},
S_d w_{12}w_{13}w_{24},
S_d w_{11}w_{22}w_{34},
S_d w_{11}w_{21}w_{33},
S_d w_{11}^2w_{12},
S_d w_{12}w_{13}w_{14}\}.
\end{align*}
By the first tangency equation (\ref{tangency_eqns}), we may study tangency arcs of invariants in $P^3(M(d,d))_0^{(M(d,d), S_d)}$ by the isotypic component of their first term $\v_1$. The isotypic decomposition of $(M(d, d),\allowbreak S_d)$, see (\ref{iso_decomp_sym2}), uses four irreducible representations of $S_d$: the trivial representation $V_{(d)}$ of degree 1, the standard representation $V_{(d-1,1)}$ of degree $d-1$, the exterior square representation $V_{(d-2,1,1)}=\wedge^2 V_{(d-1,1)}$ of degree ${(d-1)(d-2)}/{2}$, and the representation $V_{(d-2,2)}$  of degree ${d(d-3)}/{2}$ (as indicated, degrees may be computed by evaluating the respective character on the identity, giving the hook length formula, see (\pref{formula:fro})).  An explicit description of the isotypic components of $(M(d,d), S_d)$ may be given as follows \cite{arjevani2020hessian}. Consider the orthogonal direct sum $M(d, d) = \bD_d \oplus \bS_d \oplus \bA_d$, where $\bD_d$ denotes the space of $d\times d$ diagonal  matrices, $\bA_d$ the space of $d\times d$ skew-symmetric matrices and $\bS_d$ the space of $d\times d$ symmetric  matrices with diagonal entries zero. Observe that the factors are invariant under the diagonal action of $S_d$ on $M(d, d)$. For $d\ge 4$, we have,
\begin{itemize}[leftmargin=*]
\item 		
$\mathbb{D}_{d}$ is the orthogonal $S_d$-invariant direct sum $\mathbb{D}_{d,1} \oplus  \mathbb{D}_{d,2}$,
where 
\begin{enumerate}[leftmargin=*]
	\item $\mathbb{D}_{d,1} $ is the space of diagonal matrices with all entries equal and is naturally isomorphic to $V_{(d)}$. 
	\item $\mathbb{D}_{d,2}$ is the $(d-1)$-dimensional space of diagonal matrices with diagonal entries summing to zero and is naturally isomorphic to $V_{(d-1,1)}$.
\end{enumerate}
In particular, the isotypic decomposition of $\mathbb{D}_d$ is $V_{(d)} \oplus V_{(d-1, 1)}$ (isomorphic to $(\RR^d, S_d)$).

\item $\mathbb{A}_{d}$ is the orthogonal $S_d$-invariant direct sum $\mathbb{A}_{d,1} \oplus  \mathbb{A}_{d,2}$, where
\begin{enumerate}
\item $\mathbb{A}_{d,1}$ is the $(d-1)$-dimensional space of matrices $[a_{ij}]$ for which there exists $(x_1,\cdots,x_d) \in H_{d-1}$ such that for all $i,j \in \ibr{d}$,  $a_{ij} = x_i - x_j$, 
\item  $\mathbb{A}_{d,2}$ consists of all skew-symmetric matrices with row sums zero.
\end{enumerate}

As representations, $(\mathbb{A}_{d,1},S_d)$ is isomorphic to $V_{(d-1,1)}$ and $(\mathbb{A}_{d,2},S_d)$ is isomorphic to $V_{(d-2,1,1)}$.
In particular, the isotypic decomposition of $(\mathbb{A}_d,S_d)$ is $V_{(d-1,1)}\oplus V_{(d-2,1,1)}$.  

\item $\mathbb{S}_d$ is the orthogonal $S_d$-invariant direct sum $\mathbb{S}_{1,d} \oplus  \mathbb{S}_{2,d}\oplus \mathbb{S}_{3,d}$, where
\begin{enumerate}
	\item $\mathbb{S}_{d, 1}$ is the $1$-dimensional space of symmetric matrices with diagonal entries zero and all off diagonal entries equal.
	\item $\mathbb{S}_{d, 2}$ is the $(d-1)$-dimensional space of matrices $[a_{ij}]\in \mathbb{S}_d$ for which there exists $(x_1,\cdots,x_d) \in H_{d-1}$ such that
	for all $i,j \in \ibr{d}$, $i \ne j$,  $a_{ij} = x_i + x_j$.
	\item $\mathbb{S}_{d, 3}$ consists of all symmetric matrices in $\mathbb{S}_d$ with all row (equivalently, column) sums zero.
	\item $\text{dim}(\mathbb{S}_{d, 3}) = \frac{d(d-3)}{2}$.
\end{enumerate}
The representations $(\mathbb{S}_{d,i},S_d)$ are irreducible, $i \in \ibr{3}$:  $(\mathbb{S}_{d,1},S_d)$ is isomorphic to the trivial representation,
$(\mathbb{S}_{d,2})$ is isomorphic to the standard representation and
$(\mathbb{S}_{d,3},S_d)$ is isomorphic to the $V_{(d-2,2)}$.
\end{itemize}

We collect subrepresentations constituting $\bD_d, \bS_d$ and $\bA_d$ by their isomorphism type and examine the isotropy of $\gamma(t) = \sum_{i=1}^\infty t^i \v_i$ by the isotypic component of $\v_1$.\\	

\paragraph{The isotypic component $V_{(d)}$.} 
We have $V_{(d)}= M(d,d)^{\Delta S_d} = \bD_{d,1} \oplus \bS_{d,1}$. Being a fixed point space, we may restrict the differential  and so construct a tangency arc lying in $V_{(d)}$. This case accounts for the occurrence of  $\Delta S_d$ spurious minima studied in \cite{arjevani2023hidden,arjevani2020hessian,arjevani2021analytic}.\\

\paragraph{The isotypic component $V_{(d-2,1,1)}$.}
The single subrepresentation isomorphic to the exterior square representation is given by
\begin{equation} \label{eqn:x_dec}
V_{(d-2,1,1)} =  \bA_{d,2}.
\end{equation}
For the construction of analytic tangency arcs, a result analogous to \propC~requires the forth-order homogeneous component, see \secsymres. Indeed, cubic invariants all vanish on $V_{(d-2,1,1)}$
(cf. \cite[Theorem 3.6]{arjevanifield2022equivariant}). This may be shown by the method described in (\ref{sec: stab chars}) giving
\begin{align*}
\symp^3(V_{(d-2,1,1)}) &=
V_{(d-6,1,1,1,1,1,1)} \oplus V_{(d-6,2,2,1,1)} \oplus V_{(d-6,3, 3)}\\
&\oplus V_{(d-5, 2, 1, 1, 1)} \oplus V_{(d-5, 2, 2, 1)}
\oplus 2 V_{(d-5, 3, 1, 1)} \oplus V_{(d-5, 3, 2)}\\
&\oplus V_{(d-5, 4, 1)} \oplus V_{(d-4, 2, 1, 1)}^{\oplus 4}
\oplus V_{(d-4, 3, 1)}^{\oplus 5} \oplus V_{(d-4, 4)} \oplus V_{(d-3, 1, 1, 1)}^{\oplus 4}\\
& \oplus V_{(d-3, 2, 1)}^{\oplus 5} \oplus V_{(d-3, 3)}^{\oplus 3}\oplus V_{(d-2, 1, 1)}^{\oplus 6} \oplus V_{(d-2, 2)}^{\oplus 2} \oplus V_{(d-1, 1)}^{\oplus 2}.
\end{align*}
In particular, no copy of the trivial representation exists. 

We describe several isotropy groups giving one-dimensional fixed point spaces, referred to as \emph{axes of symmetry}. 
Relative to the hierarchy $(S_k\times S_1^k),k=1,2,\dots$, the relevant isotropy group is $\Delta (S_{d-2}\times S_1^2)$. Indeed, $V_{(d-2,1,1)} \cap M(d,d)^{\Delta S_d} = V_{(d-2,1,1)} \cap M(d,d)^{\Delta (S_{d-1}\times S_1)} = 0$, but
$\dim\prn{V_{(d-2,1,1)} \cap M(d,d)^{\Delta (S_{d-2}\times S_1^2)}} = 1$ with the latter given by
\begin{align}\label{eqn:x_struc}
V_{(d-2,1,1)} \cap M(d,d)^{\Delta (S_{d-2}\times S_1^2)}=	\begin{bmatrix} 
		0_{d-2}& \frac{-\alpha}{d-2}\bones_{d-2}& & \frac{\alpha}{d-2}\bones_{d-2}\\
		\frac{\alpha}{d-2}\bones_{d-2}^\top& 0&-\alpha\\
		\frac{-\alpha}{d-2}\bones_{d-2}^\top& \alpha&0
	\end{bmatrix}.
\end{align}
Additional isotropy groups are represented by the diagonal action of $S_{d-3}\times \ZZ_3 \le S_d$ on $V_{(d-2,1,1)}$, giving
\begin{align}
\left[\begin{array}{c|ccccc}
0_{d-3} \\\hline  &  0 & \alpha & - \alpha\\    &   - \alpha & 0 & \alpha\\     &   \alpha & - \alpha & 0&
\end{array}\right],
\end{align}
and $S_{d-4}\times \ZZ_4 \le S_d$ giving
\begin{align}
\left[\begin{array}{c|ccccccccc}
0_{d-4} \\\hline 
  & 0 & \alpha & 0 & - \alpha\\& - \alpha & 0 & \alpha & 0\\  &0& - \alpha & 0 & \alpha\\  &  \alpha & 0 & - \alpha & 0
\end{array}\right].
\end{align}
The two examples generalize by considering different types of cycles (e.g., by splitting factors, see \secsymres). Thus, an example in $M(7,7)$ exhibiting a somewhat more intricate structure is given by $\tri{(1, 2, 3, 4, 5, 6, 7),\allowbreak (1, 6, 2)(4, 5, 7)}$
\begin{align}
\left[\begin{matrix}0 & \alpha & \alpha & - \alpha & \alpha & - \alpha & - \alpha\\- \alpha & 0 & \alpha & \alpha & - \alpha & \alpha & - \alpha\\- \alpha & - \alpha & 0 & \alpha & \alpha & - \alpha & \alpha\\\alpha & - \alpha & - \alpha & 0 & \alpha & \alpha & - \alpha\\- \alpha & \alpha & - \alpha & - \alpha & 0 & \alpha & \alpha\\\alpha & - \alpha & \alpha & - \alpha & - \alpha & 0 & \alpha\\\alpha & \alpha & - \alpha & \alpha & - \alpha & - \alpha & 0\end{matrix}\right].
\end{align}\\

\paragraph{The isotypic component $V_{(d-2,2)}$.} The  subrepresentation in the isotopic decomposition $(M(d, d),S_d)$ isomorphic to  $V_{(d-2,2)}$ is given by
\begin{equation} \label{eqn:y_dec}
	V_{(d-2,2)} =  \bS_{d,3}.
\end{equation}
Referring again to the hierarchy $\Delta (S_k\times S_1^k),k=1,2,\dots$, the relevant isotropy group is $\Delta (S_{d-2}\times S_1^2)$ as  $V_{(d-2,2)} \cap M(d,d)^{\Delta S_d} = V_{(d-2,2)} \cap M(d,d)^{\Delta (S_{d-1}\times S_1)} = 0$, but
$\dim\prn{V_{(d-2,2)} \cap M(d,d)^{\Delta (S_{d-2}\times S_1^2)}} = 1$ with the latter given by 
\begin{align*}
	\begin{bmatrix}
		\frac{2\alpha}{(d-2)(d-3)}(\bones_{d-2}\bones_{d-2}^\top - I_{d-2})& \frac{-\alpha}{d-2}\bones_{d-2}& & \frac{-\alpha}{d-2}\bones_{d-2}\\
		\frac{-\alpha}{d-2}\bones_{d-2}^\top& 0&\alpha\\
		\frac{-\alpha}{d-2}\bones_{d-2}^\top& \alpha&0
	\end{bmatrix}.
\end{align*}\\

\paragraph{The isotypic component $V_{(d-1,1)}$.}
The three subrepresentations isomorphic to $V_{(d-1,1)}$ are 
\begin{equation} \label{eqn:std_dec}
	V_{(d-1,1)} = \bD_{d,2} \oplus \bS_{d,2} \oplus \bA_{d,1}.
\end{equation}
Referring to the hierarchy  $\Delta (S_k\times S_1^k),k=1,2,\dots$, the relevant isotropy group is $\Delta (S_{d-1}\times S_1)$ as $V^{\oplus 3}_{(d-1,1)} \cap M(d,d)^{\Delta S_d} = 0$ but
$\dim\prn{V_{(d-1,1)} \cap M(d,d)^{\Delta (S_{d-1}\times S_1)}} = 3$ with 
\begin{align}\label{std_struc}
	\bD_{d,2} \cap M(d,d)^{\Delta S_{d-1}} &= \frac{-\alpha}{d-1}I_{d-1}\oplus [\alpha],\nonumber\\
	\bS_{d,2} \cap M(d,d)^{\Delta S_{d-1}} &= 
	\begin{bmatrix}
		\frac{-2}{d-2}\alpha(\bones_{d-1} \bones_{d-1}^\top - I_{d-1})& \alpha\bones_{d-1},\\
		\alpha\bones_{d-1}^\top& 0
	\end{bmatrix},
	\\
	\bA_{d,1} \cap M(d,d)^{\Delta S_{d-1}} &= 
	\begin{bmatrix}
		0_d& -\alpha\bones_{d-1}\\
		\alpha\bones_{d-1}^\top& 0
	\end{bmatrix}.\nonumber
\end{align}
Additional isotropy groups giving three dimensional fixed point subspace are given by $\tri{(1,2,\dots,d-k), (d-k+1,\dots,d)}$ for different choice of $k$. For example, the fixed point space corresponding to $k=3$ and $d=7$ is
\begin{align*}
\left[\begin{matrix}- \frac{4 \alpha_{1}}{3} & - \frac{8 \alpha_{2}}{3} & - \frac{8 \alpha_{2}}{3} & - \frac{\alpha_{2}}{3} + \frac{7 \alpha_{3}}{4} & - \frac{\alpha_{2}}{3} + \frac{7 \alpha_{3}}{4} & - \frac{\alpha_{2}}{3} + \frac{7 \alpha_{3}}{4} & - \frac{\alpha_{2}}{3} + \frac{7 \alpha_{3}}{4}\\- \frac{8 \alpha_{2}}{3} & - \frac{4 \alpha_{1}}{3} & - \frac{8 \alpha_{2}}{3} & - \frac{\alpha_{2}}{3} + \frac{7 \alpha_{3}}{4} & - \frac{\alpha_{2}}{3} + \frac{7 \alpha_{3}}{4} & - \frac{\alpha_{2}}{3} + \frac{7 \alpha_{3}}{4} & - \frac{\alpha_{2}}{3} + \frac{7 \alpha_{3}}{4}\\- \frac{8 \alpha_{2}}{3} & - \frac{8 \alpha_{2}}{3} & - \frac{4 \alpha_{1}}{3} & - \frac{\alpha_{2}}{3} + \frac{7 \alpha_{3}}{4} & - \frac{\alpha_{2}}{3} + \frac{7 \alpha_{3}}{4} & - \frac{\alpha_{2}}{3} + \frac{7 \alpha_{3}}{4} & - \frac{\alpha_{2}}{3} + \frac{7 \alpha_{3}}{4}\\- \frac{\alpha_{2}}{3} - \frac{7 \alpha_{3}}{4} & - \frac{\alpha_{2}}{3} - \frac{7 \alpha_{3}}{4} & - \frac{\alpha_{2}}{3} - \frac{7 \alpha_{3}}{4} & \alpha_{1} & 2 \alpha_{2} & 2 \alpha_{2} & 2 \alpha_{2}\\- \frac{\alpha_{2}}{3} - \frac{7 \alpha_{3}}{4} & - \frac{\alpha_{2}}{3} - \frac{7 \alpha_{3}}{4} & - \frac{\alpha_{2}}{3} - \frac{7 \alpha_{3}}{4} & 2 \alpha_{2} & \alpha_{1} & 2 \alpha_{2} & 2 \alpha_{2}\\- \frac{\alpha_{2}}{3} - \frac{7 \alpha_{3}}{4} & - \frac{\alpha_{2}}{3} - \frac{7 \alpha_{3}}{4} & - \frac{\alpha_{2}}{3} - \frac{7 \alpha_{3}}{4} & 2 \alpha_{2} & 2 \alpha_{2} & \alpha_{1} & 2 \alpha_{2}\\- \frac{\alpha_{2}}{3} - \frac{7 \alpha_{3}}{4} & - \frac{\alpha_{2}}{3} - \frac{7 \alpha_{3}}{4} & - \frac{\alpha_{2}}{3} - \frac{7 \alpha_{3}}{4} & 2 \alpha_{2} & 2 \alpha_{2} & 2 \alpha_{2} & \alpha_{1}\end{matrix}\right].
\end{align*}

\phantom{}\\
\noindent \textbf{Acknowledgements}\\
We thank Noa and Tom. The research was supported by the Israel
Science Foundation (grant No. 724/22).

\bibliographystyle{ieeetr}
\bibliography{bib}

\end{document}